%% file: main.tex
\documentclass[letterpaper]{article} % DO NOT CHANGE THIS
\usepackage[draft]{aaai2026}  % DO NOT CHANGE THIS
\usepackage{times}  % DO NOT CHANGE THIS
\usepackage{helvet}  % DO NOT CHANGE THIS
\usepackage{courier}  % DO NOT CHANGE THIS
\usepackage[hyphens]{url}  % DO NOT CHANGE THIS
\usepackage{graphicx} % DO NOT CHANGE THIS
\usepackage{natbib}  % DO NOT CHANGE THIS AND DO NOT ADD ANY OPTIONS TO IT
\usepackage{caption} % DO NOT CHANGE THIS AND DO NOT ADD ANY OPTIONS TO IT
\usepackage{algorithmic}
\usepackage{booktabs}
\usepackage{multirow}
\usepackage{siunitx}
\usepackage{makecell}
\usepackage{amsmath}
\usepackage{tabularx}
\usepackage{enumitem}
\usepackage{framed}
\usepackage{amssymb}
\usepackage[ruled,linesnumbered]{algorithm2e}
\usepackage{bm}
\usepackage{amsfonts}       % blackboard math symbols
\usepackage{nicefrac}       % compact symbols for 1/2, etc.
\usepackage{microtype}      % microtypography
\usepackage[dvipsnames]{xcolor}
\usepackage{listings}
\usepackage{subcaption}
\title{Navigating the Alpha Jungle:\\ An LLM-Powered MCTS Framework for Formulaic Factor Mining}
\author{
  Yu Shi\textsuperscript{\rm 1}, Yitong Duan\textsuperscript{\rm 1, \rm 2}, and Jian Li\textsuperscript{\rm 1}
}
\affiliations{
    \textsuperscript{\rm 1}Institute for Interdisciplinary Information Sciences, Tsinghua University\\
    \textsuperscript{\rm 2}Zhongguancun Institute of Artificial Intelligence\\
  \{shi-y23, dyt19\}@mails.tsinghua.edu.cn, lapordge@gmail.com 
}

\newcommand{\op}[1]{\mathrm{#1}}

\DontPrintSemicolon
\SetAlgoLined

\SetCommentSty{mycommentsty}
\SetKwComment{Comment}{\color{RoyalBlue}\footnotesize$\triangleright$\ }{}

\newcommand{\algphase}[1]{\tcc*{\textcolor{NavyBlue}{\textbf{#1}}}}
\newcommand{\subphase}[1]{\tcc{\textcolor{DarkOrchid}{\textit{#1}}}}

\newcommand{\crossmark}{$\times$}

\begin{document}

\maketitle

\begin{abstract}
\input{sections/Abstract}
\end{abstract}

\input{sections/Introduction}

\input{sections/Preliminary_v2}

\input{sections/Methodology_v3}

\input{sections/Experiment}

\input{sections/Related_Work}

\input{sections/Conclusion}

% \bibliography{aaai26}

\clearpage
\appendix

\setcounter{secnumdepth}{1}
\input{sections/Appendix}

\end{document}

%% file: sections/Abstract.tex
Alpha factor mining is pivotal in quantitative investment for identifying predictive signals from complex financial data. While traditional formulaic alpha mining relies on human expertise, contemporary automated methods, such as those based on genetic programming or reinforcement learning, often struggle with search inefficiency or yield alpha factors that are difficult to interpret. This paper introduces a novel framework that integrates Large Language Models (LLMs) with Monte Carlo Tree Search (MCTS) to overcome these limitations. Our framework leverages the LLM's instruction-following and reasoning capability to iteratively generate and refine symbolic alpha formulas within an MCTS-driven exploration. A key innovation is the guidance of MCTS exploration by rich, quantitative feedback from financial backtesting of each candidate factor, enabling efficient navigation of the vast search space. Furthermore, a frequent subtree avoidance mechanism is introduced to enhance search diversity and prevent formulaic homogenization, further improving performance. Experimental results on real-world stock market data demonstrate that our LLM-based framework outperforms existing methods by mining alphas with superior predictive accuracy and trading performance. The resulting formulas are also more amenable to human interpretation, establishing a more effective and efficient paradigm for formulaic alpha mining.

%% file: sections/Introduction.tex
\section{Introduction}
\label{sec:introduction}

Predicting price movements in financial markets, characterized by low signal-to-noise ratios, remains a central challenge in quantitative investment. A common strategy to enhance model predictiveness is the extraction of predictive signals, or alpha factors (hereafter simply ``alphas''), from stock data~\cite{qian2007quantitative, tulchinsky2019finding}. Current alpha factor mining methodologies broadly fall into two categories: neural network-based and formula-based. Neural approaches (e.g., FactorVAE~\cite{duan2022factorvae}, HIST~\cite{xu2021hist}, REST~\cite{xu2021rest}) implicitly construct complex alphas via deep learning, capturing intricate patterns but often suffering from a lack of interpretability. In contrast, formula-based methods aim to discover alphas represented by explicit mathematical expressions. These alpha factors are traditionally human-crafted, reflecting market insights (e.g., Fama-French factors~\cite{fama1992cross}, financial anomalies~\cite{harvey2016and,hou2020replicating}). In recent years, automated techniques have emerged, employing methods like genetic programming or reinforcement learning to discover such formulaic alphas~\cite{shi2024alphaforge,yu2023generating,zhang2020autoalpha,zhang2023openfe}.

\begin{figure}[t!]
  \vspace{-\intextsep}
  \centering
  \includegraphics[width=\linewidth]{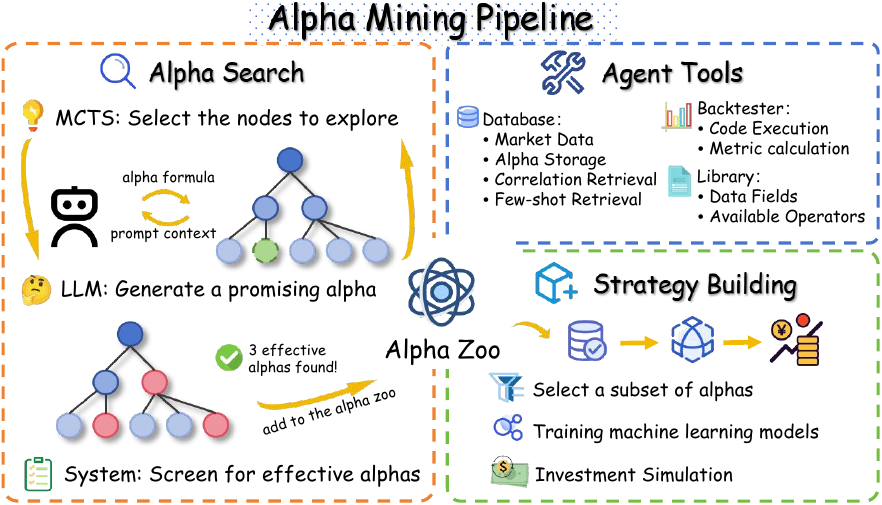}
  \caption{A high-level schematic of our proposed alpha mining pipeline. The pipeline features an iterative Alpha Search loop where an LLM, guided by MCTS, generates and refines formulas. Effective alphas are collected in an Alpha Zoo before being used in the final Strategy Building stage.}
  \label{fig:pipeline}
  \vspace{-\intextsep}
\end{figure}

Despite their promise, existing automated formulaic alpha mining approaches face significant limitations.
\textbf{First, the discovered alphas often exhibit poor interpretability.} These automated methods frequently engage in unconstrained, data-driven exploration of the vast alpha space, often without sufficient guidance from financial theory or domain expertise. Consequently, the resulting formulas can be overly complex and opaque. This lack of transparency poses considerable challenges in practical investment scenarios: it hinders practitioners' ability to understand the underlying economic rationale of a strategy, makes it difficult to attribute portfolio performance accurately, and can erode trust, thereby impeding the adoption of these alphas even if they show promise in backtests.
\textbf{Second, current methodologies often suffer from search inefficiency.} The search for a sufficient number of effective alpha factors typically requires generating and evaluating an enormous volume of candidate formulas. This exhaustive search process, while necessary due to the low signal density, inherently increases the likelihood of discovering spurious relationships and overfitting to the training data~\cite{harvey2016and}. As a result, many discovered alphas may exhibit poor generalization and deliver underwhelming out-of-sample performance.

Addressing the identified shortcomings necessitates innovative approaches. In this light, Large Language Models (LLMs) emerge as a promising direction, given their vast prior knowledge and strong reasoning capabilities which are well-suited for generating interpretable alphas—a potential demonstrated in analogous tasks like financial investment~\cite{yu2024finmem} and code generation~\cite{li2024rethinkmcts}. Drawing inspiration from advancements in LLM reasoning (e.g., Chain-of-Thought~\cite{wei2022chain}, Tree-of-Thought~\cite{yao2024tree}) and the efficacy of Monte Carlo Tree Search (MCTS)~\cite{coulom2007monte,silver2016mastering} in enhancing LLM performance on complex problems~\cite{zhang2024accessing}, we frame alpha mining as an MCTS-driven search problem. Within this framework, each node in the tree represents a candidate alpha formula, allowing for a systematic exploration and iterative refinement within the vast and complex alpha space. Figure~\ref{fig:pipeline} provides a high-level illustration of this entire pipeline.

Unlike tasks such as mathematical derivation, where evaluating the contribution of intermediate steps towards the final solution is often challenging before the derivation is complete, alpha factor mining provides fine-grained feedback on each candidate alpha formula through backtesting. We leverage this detailed feedback to guide our search. We initiate the search with a LLM-generated alpha formula, as the root node of the search tree. Then we utilize the LLM to iteratively refine and improve the formulas, expanding the tree with new, potentially superior nodes. Furthermore, to mitigate the homogeneity of generated alpha formulas, we conduct frequent subtree mining on effective alphas and explicitly instruct the LLM to avoid using the most frequent subtrees during generation. By explicitly diversifying the search away from these common motifs, our approach enhances search efficiency and improves the quality of discovered alphas.

The synergy between MCTS and LLMs has indeed shown promise in various reasoning tasks~\cite{zhao2023large,delorenzo2024make}. Our work is distinct in its specific application of this synergy to the unique challenges of formulaic alpha mining. Unlike general reasoning tasks that use MCTS to explore a set of predefined actions or have the LLM evaluate abstract states~\cite{xie2024monte,li2025enhancing,dainese2024generating}, our framework leverages the LLM as a \textit{generative prior} for symbolic alpha formulas. Crucially, the MCTS exploration is guided by rich, quantitative, and domain-specific feedback from financial backtesting performed on each candidate alpha. This iterative loop—where the LLM's generative capabilities are steered by MCTS informed by empirical financial performance—offers a distinct advantage.

The main contributions of this paper can be summarized as follows:

\begin{itemize}
    \item We propose an LLM-Powered MCTS framework for formulaic alpha mining, modeling the task as a tree search-based reasoning problem where the LLM performs multi-step formula refinement guided by detailed backtesting feedback.
    \item We design a frequent subtree avoidance method to improve search efficiency and alpha effectiveness by guiding the LLM to explore less common yet potentially effective formula structures.
    \item We conduct a series of experiments to demonstrate the effectiveness of our proposed framework. The alphas mined by our framework achieve superior prediction performance while maintaining good interpretability, compared to those from other methods.
\end{itemize}

%% file: sections/Preliminary_v2.tex
\section{Preliminary}

\subsection{Alpha Factor Mining}

We consider a financial market with $n$ stocks observed over $T$ trading days. For each stock $i \in \{1, \dots, n\}$ and day $t \in \{1, \dots, T\}$, its state is described by a feature vector $\boldsymbol{x}_{i,t} \in \mathbb{R}^m$. Raw features include daily open, high, low, close prices (OHLC), trading volume, and Volume-Weighted Average Price (VWAP). The complete market history is a tensor $\boldsymbol{X} \in \mathbb{R}^{T \times n \times m}$. Correspondingly, future returns are organized in a matrix $\boldsymbol{Y} \in \mathbb{R}^{T \times n}$, where $y_{i,t}$ is the realized future return for stock $i$ subsequent to day $t$. To capture temporal patterns, we use a lookback window of length $\tau$. An \textit{alpha factor}, $f$, maps the historical feature data for this window, $\boldsymbol{X}_{t-\tau+1:t} = \{\boldsymbol{X}_s \mid t-\tau < s \le t\}$, to a vector of predictive scores $\boldsymbol{v}_t = f(\boldsymbol{X}_{t-\tau+1:t}) \in \mathbb{R}^n$. Each $v_{i,t}$ represents the alpha's assessment of stock $i$'s future return.

Alpha factor mining aims to discover a diverse set of $K$ alphas, $\mathcal{F} = \{f_1, \dots, f_K\}$, by searching within the vast space of all possible alpha factors, denoted as $\mathcal{A}$. The outputs of these individual alphas, $\{\boldsymbol{v}_{k,t} = f_k(\boldsymbol{X}_{t-\tau+1:t})\}_{k=1}^K$, are typically aggregated by a combination model, $g$, into a composite alpha vector $\boldsymbol{z}_t = g(\{\boldsymbol{v}_{k,t}\}_{k=1}^K ; \boldsymbol{\theta}_g)$, where $\boldsymbol{\theta}_g$ are model parameters.
The quality of this composite signal is evaluated using a predefined performance metric, $\mathcal{L}$ (e.g., Information Coefficient). Optimal parameters for the combination model, $\boldsymbol{\theta}_g^*$, are learned by maximizing this metric:
\begin{equation}
\boldsymbol{\theta}_g^*(\mathcal{F}) = \arg\max_{\boldsymbol{\theta}_g} \mathcal{L}(g(\{\boldsymbol{v}_{k,t}\}_{k=1}^K; \boldsymbol{\theta}_g), \boldsymbol{Y}).
\end{equation}
Let $g^*(\mathcal{F})$ be the combination model with parameters $\boldsymbol{\theta}_g^*(\mathcal{F})$. The overarching goal is to find an optimal set of alpha factors $\mathcal{F}^* \subset \mathcal{A}$ that maximizes the performance of this optimally combined signal: $\mathcal{F}^* = \arg\max_{\mathcal{F} \subset \mathcal{A}} \mathcal{L}(g^*(\mathcal{F}), \boldsymbol{Y}).$
This constitutes a challenging bilevel optimization problem due to the immense search space $\mathcal{A}$ and the complex interactions between alpha factors.

\begin{figure*}[t!]
    \centering
    \includegraphics[width=1.0\textwidth]{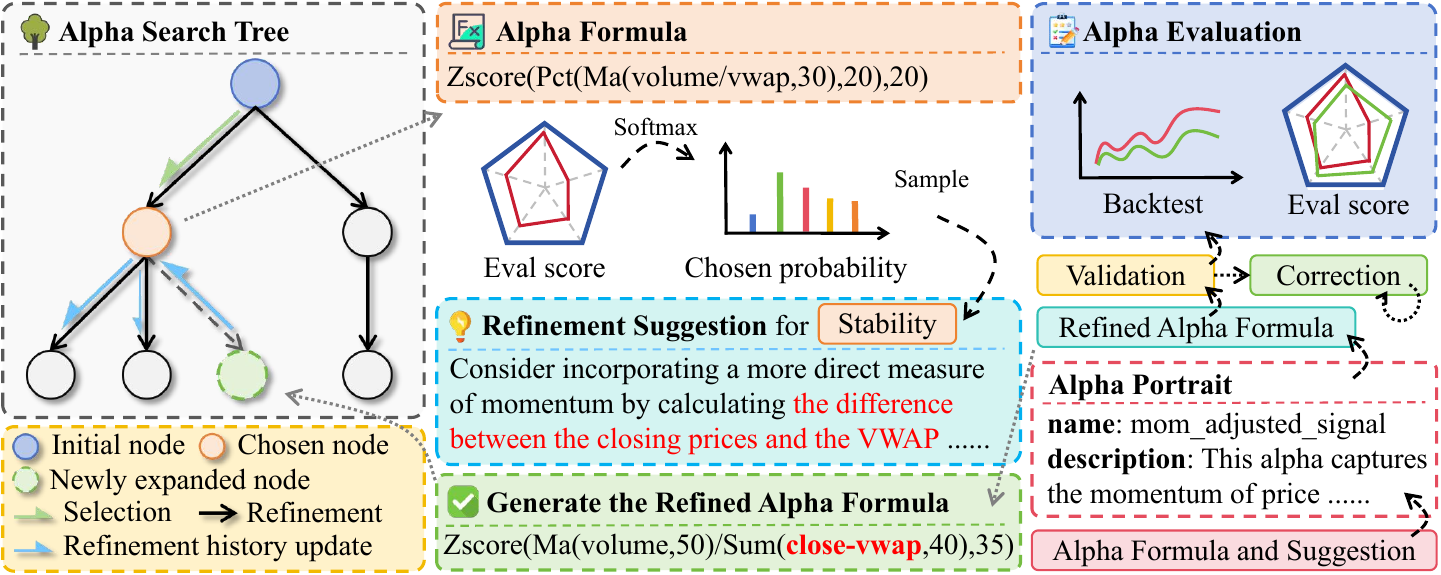}
    \caption{Overview of our LLM-powered MCTS framework. The process begins with node selection via UCT. A refinement dimension is then chosen based on the node's multi-dimensional evaluation scores. The LLM first proposes a conceptual refinement suggestion for that dimension and then translates it into a concrete formula. The new formula is backtested, and its performance results are used to expand the tree with a new node.}
    \label{fig:overview}
\end{figure*}

\subsection{Formulaic Alpha}

In this work, we focus on \textit{formulaic alphas}: alpha factors defined by mathematical expressions. These expressions are constructed from operators and operands. Operands typically include raw input features (e.g., $\text{close}_{i,t}$) and numerical constants. Operators apply mathematical transformations; for example, time-series operators can be used to construct an alpha like $\text{Ma}(\text{close}, 5) - \text{Ma}(\text{close}, 20)$. This specific alpha captures a price trend by contrasting short-term with long-term moving averages of closing prices. A complete list of available operators is provided in Appendix~\ref{sec:appendix_operators}. Formulaic alphas are naturally represented as expression trees (leaf nodes: operands; internal nodes: operators), making their structured yet flexible nature amenable to the automated mining techniques central to our framework.

%% file: sections/Methodology_v3.tex
\section{Methodology}
\label{sec:methodology}

Our proposed alpha mining framework integrates LLMs with MCTS to automate the discovery and refinement of alpha factors. The framework's objective is to search within the vast space of possible alpha formulas, denoted as $\mathcal{A}$, to find a set of high-performing alphas. Figure~\ref{fig:overview} provides a conceptual illustration. The core iterative process involves: (1) selecting a promising node (alpha formula) using the Upper Confidence Bound for Trees (UCT) criterion \cite{kocsis2006bandit}; (2) expanding this node by having the LLM generate a refined alpha, guided by performance feedback on specific evaluation dimensions; and (3) evaluating the new alpha via backtesting, with results forming a new node in the search tree $\mathcal{T}$ (see Appendix~\ref{sec:appendix_case_study} for an illustrative case). The LLM's role is twofold: first, to propose targeted refinement suggestions, and second, to translate these suggestions into a concrete alpha formula $f \in \mathcal{A}$. Iteratively, high-performing alphas that satisfy a set of predefined criteria $\mathcal{C}$ (e.g., $\text{IC} > 0.02$) are collected into an effective alpha repository, $\mathcal{F}_{zoo}$.

\subsection{Selection}

The selection step in our MCTS framework navigates the exploration-exploitation trade-off. Each node $s \in \mathcal{T}$ represents an alpha, characterized by its formula $f_s$ and refinement history. An action $a$ corresponds to a specific refinement applied to $s$. Each state-action pair $(s, a)$ maintains a quality value $Q(s, a)$, representing the maximum reward (alpha score) observed in the subtree rooted at the child node resulting from this action. We employ the UCT criterion to select the optimal action $a^*$:
\begin{equation}
a^* = \arg\max_{a \in A(s)} \left( Q(s, a) + c\sqrt{\frac{\ln (N_s)}{N_{s'}}} \right)
\label{eq:uct}
\end{equation}
where $A(s)$ is the set of existing actions from state $s$, $N_s$ is the visit count of the parent state $s$, $N_{s'}$ is the visit count of the child state $s' = \text{child}(s, a)$, and $c$ is the exploration weight.

Unlike standard MCTS, which expands only leaf nodes, our approach allows any node to be selected for expansion. This is crucial for iteratively refining promising, but not terminal, alpha ideas. To enable this, we augment the action space $A(s)$ at any internal node $s$ with a ``virtual'' expansion action, $a_{e}$. The selection process thus considers the full action set $A(s) \cup \{a_{e}\}$. The UCT score for this virtual action is computed by adapting Equation~\ref{eq:uct}, where we define a virtual visit count for the prospective new node $s'_{e}$ as $N_{s'_{e}} = 1 + |C(s)|$, with $C(s)$ being the set of existing children of $s$. If $a_{e}$ is selected, node $s$ is chosen for expansion. This mechanism ensures that promising, non-leaf nodes can be further refined.

\subsection{Expansion}

Upon selecting a node $s$ for expansion, a new, refined alpha factor $f_{new}$ is generated. This process is structured to enhance the LLM's effectiveness and the quality of refinements.

\textbf{Dimension-Targeted Refinement Suggestion.} Each node $s$ is associated with a multi-dimensional evaluation score vector $\boldsymbol{E}_s = [e_1, \dots, e_q] \in [0, e_{\text{max}}]^q$. To guide refinement towards areas of weakness while maintaining explorative diversity, we stochastically select a target dimension $i^*$ for improvement. The probability of choosing dimension $i$ is defined as:
\begin{equation}
P(i^*=i|s) = \text{Softmax}\left((e_{\text{max}} \cdot \boldsymbol{1}_q - \boldsymbol{E}_s)/T\right)_i
\label{eq:refinement_sampling_criterion}
\end{equation}
where $\boldsymbol{1}_q$ is a $q$-dimensional vector of ones and $T$ is a temperature parameter. This strategy prioritizes dimensions with lower scores. Once a dimension $i^*$ is selected, the LLM generates a textual refinement suggestion $d_{s, i^*}$ aimed at improving performance on that dimension. This is framed as a few-shot learning task, where the context includes effective alphas from $\mathcal{F}_{zoo}$ (see Appendix~\ref{sec:appendix_method_details}).

\textbf{Alpha Formula Generation and Validation.} Following the targeted suggestion, we employ a two-step generation process. First, the LLM articulates the refined conceptual hypothesis, which is then used to prompt the generation of the concrete formula. This process, ensuring the formula aligns with a clear investment rationale, can be formalized as:
\begin{align}
d_{s, i^*} &\sim p_{\text{LLM}}(\cdot | s, i^*, \mathcal{F}_{zoo}) \label{eq:llm_suggestion} \\
f_{new} &\sim p_{\text{LLM}}(\cdot | d_{s, i^*}, f_s) \label{eq:llm_generation}
\end{align}
The generated formula $f_{new}$ undergoes an automated validation check, $\text{IsValid}(f_{new})$. If invalid, feedback is provided to the LLM for iterative correction. A valid formula and its evaluation results constitute the new node $s_{new}$ in the MCTS tree.

\subsection{Multi-Dimensional Alpha Evaluation}
\label{sec:evaluation}

The evaluation of a candidate alpha $f$ is performed directly via backtesting, bypassing the simulation phase of traditional MCTS. A key challenge is the evolving nature of the effective alpha repository $\mathcal{F}_{zoo}$, which progressively raises the bar for new alphas. To address this, we employ a relative ranking approach. The rank of $f$ against the repository for a given metric $m$ is:
\begin{equation}
R(f, m, \mathcal{F}_{zoo}) = \frac{1}{|\mathcal{F}_{zoo}|}\sum_{f' \in \mathcal{F}_{zoo}} \mathbb{I}(m(f) < m(f'))
\label{eq:relative_rank}
\end{equation}
where $\mathbb{I}(\cdot)$ is the indicator function. This provides an adaptive evaluation criterion, avoiding fixed thresholds that may be too stringent early on or too lenient later.

To provide granular feedback for refinement, we evaluate $f$ across a set of $q$ dimensions $\mathcal{D}$, exemplified here by \textbf{Effectiveness}, \textbf{Stability}, \textbf{Turnover}, \textbf{Diversity}, and \textbf{Overfitting Risk} (detailed in Appendix~\ref{sec:appendix_method_details}). For each dimension $D_i \in \mathcal{D} \setminus \{\text{Overfitting Risk}\}$, we compute a score $e_i$ based on its percentile rank using a corresponding metric $m_i$:
\begin{equation}
e_i(f) = 1 - R(f, m_i, \mathcal{F}_{zoo})
\label{eq:dim_score}
\end{equation}
The assessment of \textbf{Overfitting Risk}, denoted $e_{\text{overfit}}$, is distinct. We leverage an LLM that analyzes the formula $f$ and its refinement history $H(s)$, providing a qualitative judgment: $e_{\text{overfit}} = \text{LLM}_{\text{eval}}(f, H(s))$ (see Appendix~\ref{sec:alpha_overfitting_risk_prompt} for the specific prompt). The overall alpha score, which serves as the reward signal for MCTS, is the aggregate of these dimensional scores:
\begin{equation}
S(f) = \frac{1}{|\mathcal{D}|} \sum_{i=1}^{q} e_i(f)
\label{eq:overall_score}
\end{equation}
This score $S(f)$ is used to update the Q-values in the search tree.

\subsection{Backpropagation}

During backpropagation, the reward $S(f_{new})$ of the newly evaluated alpha at node $s_{new}$ updates the statistics of all nodes along the path from the root to its parent. For each ancestor node $s_k$ on this path, leading to its child $s_{k+1}$ via action $a_k$, we perform the following updates:
\begin{align}
N_{s_k} &\leftarrow N_{s_k} + 1 \\
Q(s_k, a_k) &\leftarrow \max(Q(s_k, a_k), S(f_{new}))
\label{eq:q_update}
\end{align}
This ensures that the value of a path reflects the best outcome discovered within its entire subtree. Crucially, to enhance the quality of subsequent refinement suggestions and the accuracy of overfitting assessment, the LLM is provided with rich contextual information: the refinement history of the current node's parent, children, and siblings. This allows the LLM to analyze the refinement trajectory and avoid redundant suggestions.

\begin{figure}[t!]
  \vspace{-\intextsep}
  \centering
  \includegraphics[width=\linewidth]{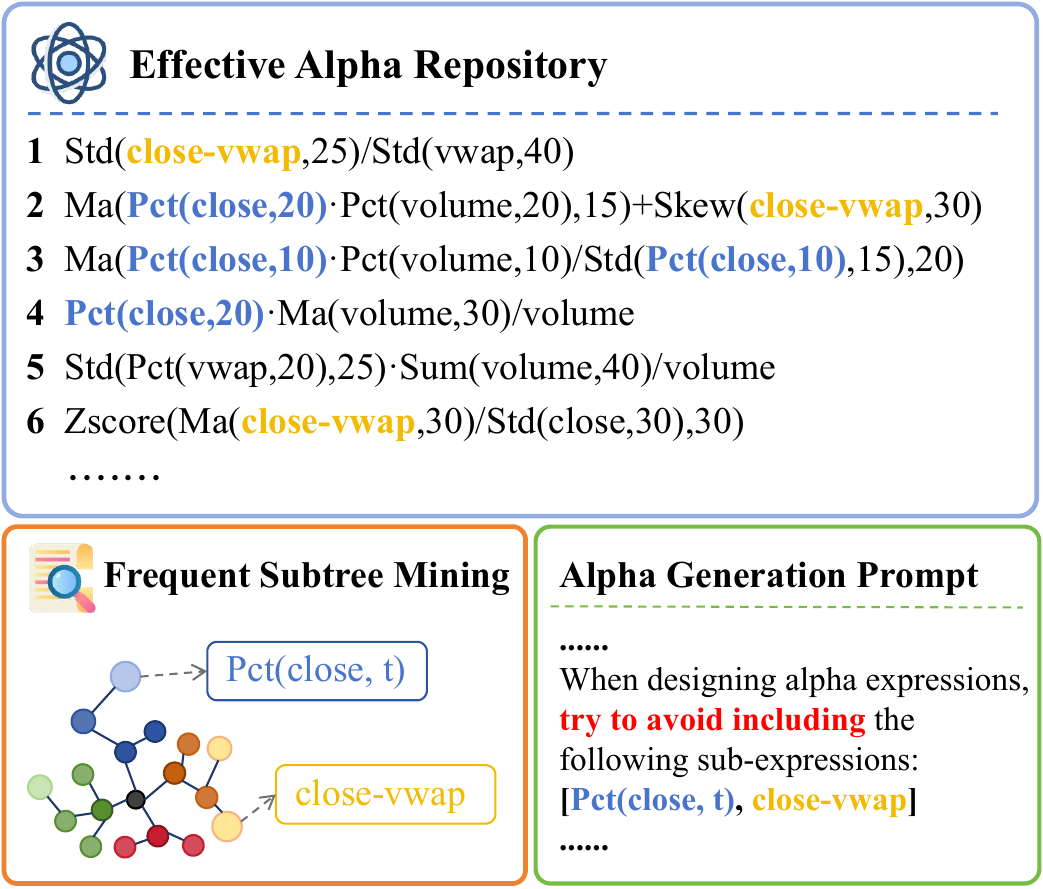}
  \caption{Illustration of the Frequent Subtree Avoidance (FSA). The set of effective alphas from the Alpha Repository is mined for frequent subtrees. The most frequent ones are identified, and the LLM is subsequently instructed to avoid generating new formulas containing these common structural motifs.}
  \label{fig:freq_subtree_desp}
  \vspace{-\intextsep}
\end{figure}

\subsection{Frequent Subtree Avoidance}

To mitigate alpha formula homogenization and prevent the over-exploitation of common motifs, we introduce Frequent Subtree Avoidance (FSA), inspired by the concept of ``root genes'' in AutoAlpha \cite{zhang2020autoalpha}. We define a \textit{root gene} as a subtree in an alpha's expression tree whose leaves are exclusively raw input features (e.g., `close', `high'). To focus on structure, we use an operator $\text{Abs}(\cdot)$ that abstracts away concrete parameter values from an expression tree. For instance, $\text{Abs}(\text{Ma}(\text{vwap}, 20))$ becomes $\text{Ma}(\text{vwap}, t)$. The set of abstracted root genes for an alpha $f$ is denoted $\bar{\mathcal{G}}(f)$.

The FSA mechanism operates by first identifying frequent closed root genes from the repository $\mathcal{F}_{zoo}$. A root gene is ``closed'' if none of its immediate supertrees share the same support count, which helps identify maximal common patterns. The support for an abstracted root gene $\bar{g}$ is:
\begin{equation}
\text{Support}(\bar{g}) = \frac{1}{|\mathcal{F}_{zoo}|} \sum_{f' \in \mathcal{F}_{zoo}} \mathbb{I}(\bar{g} \subseteq \bar{\mathcal{G}}(f'))
\label{eq:support}
\end{equation}
We select the top-$k$ most frequent closed root genes to form a set of forbidden structures, $\mathcal{G}_{\text{forbidden}}$. During generation (Eq.~\ref{eq:llm_generation}), the LLM is constrained to produce a new formula $f_{new}$ that avoids these motifs:
\begin{equation}
\text{Constraint: } \bar{\mathcal{G}}(f_{new}) \cap \mathcal{G}_{\text{forbidden}} = \emptyset
\label{eq:fsa_constraint}
\end{equation}
FSA acts as a regularization on the generation process. By discouraging common structures, it guides the MCTS search towards more structurally diverse and potentially novel regions of the alpha space $\mathcal{A}$, enabling a more efficient exploration, as demonstrated in Section~\ref{sec:ablation_study}.

%% file: sections/Experiment.tex
\section{Experiment}
\label{sec:experiment}

We evaluate our proposed framework on real-world stock market data, addressing the following research questions (RQs):\\
\textbf{Q1}: How does our approach compare to baselines in predictive performance? \\
\textbf{Q2}: Are MCTS and Frequent Subtree Avoidance effective components within our framework? \\
\textbf{Q3}: How does the interpretability of alpha formulas mined by our method compare to others?

\subsection{Experiment Settings}

\paragraph{Data} Our experiments are conducted on the Chinese A-shares market. To ensure comprehensive market representation, our experiments separately target two stock pools: the CSI 300 Index (large-cap, liquid stocks) and the CSI 1000 Index (small to mid-cap stocks). We define two distinct prediction targets: the 10-day return and the 30-day return of the stocks, with buying and selling executed at the closing price. The dataset is split chronologically into training (2011/01/01--2020/12/31) and testing (2021/01/01--2024/11/30) periods.

\paragraph{Baselines for Comparison} We compare our framework with several formulaic alpha mining methods. \textbf{DSO} (Deep Symbolic Optimization) \cite{landajuela2022unified} is a deep learning framework for symbolic optimization. \textbf{GP} employs genetic programming for alpha mining. \textbf{AlphaGen} \cite{yu2023generating} is a reinforcement learning framework for formulaic alpha mining. \textbf{AlphaForge} \cite{shi2024alphaforge} features a generative-predictive architecture; to ensure a fair comparison of generative capabilities, we use only its alpha mining network. Among LLM-based approaches, \textbf{CoT} (Chain-of-Thought) \cite{wei2022chain} prompts LLMs for step-by-step reasoning to directly generate alpha factors. \textbf{ToT} (Tree-of-Thought) \cite{yao2024tree} enables LLMs to explore diverse reasoning paths in a tree structure. Lastly, \textbf{FAMA} \cite{li2024can} leverages LLMs with in-context examples to diversify formulas and a ``chain-of-experience'' to learn from past successes.

We use OpenAI's GPT4.1 model as the LLM for both our method and the LLM-based baselines. Furthermore, we evaluate the performance of our framework when using different LLM backbones, as detailed in Appendix~\ref{sec:llm_sensitivity}. Regarding the potential data leakage issue in LLMs, please refer to the discussion in Appendix~\ref{sec:data_leakage_in_llm}.

To ensure a fair and rigorous comparison of algorithmic efficiency, we benchmark all methods based on a controlled ``search count'' (i.e., the number of unique alpha formulas generated and evaluated). This metric normalizes for the vast differences in computational cost per generation step across methods (e.g., a single LLM call vs. a GP mutation) and provides a direct measure of search space exploration efficiency. This approach is well-suited because our framework and all baselines inherently involve a distinct search process, where each iteration yields a new candidate alpha formula. 
For LLM-based methods (including ours), we report the best performance achieved with search counts of 1,000, 2,000, or 3,000.
For other methods, the search count is incrementally increased from a small value until performance converges, capped at 600,000 (200$\times$ the LLM-based methods' maximum).
This experimental design facilitates two key comparisons: first, it allows for an equitable assessment of our framework against other LLM-based methods under similar, well-defined computational budgets; second, it enables a robust comparison of search efficiency against other non-LLM-based methods.

\begin{figure}[t]
  \vspace{-\intextsep}
  \centering
  \includegraphics[width=1.0\columnwidth]{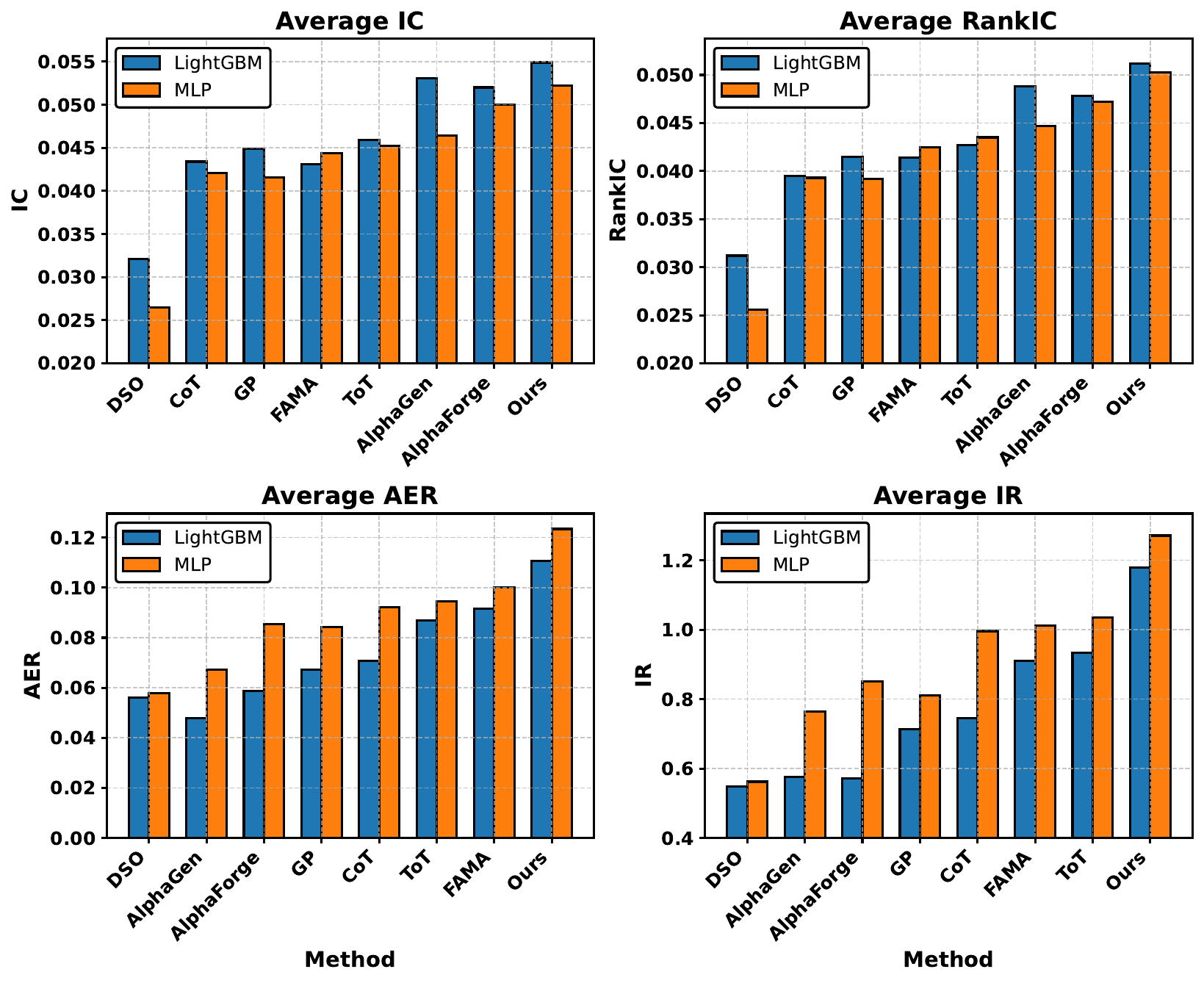}
  \caption{The average predictive performance of LightGBM and MLP models trained on alphas mined by different methods.}
  \label{fig:method_performance}
  \vspace{-\intextsep}
\end{figure}

\subsection{Experiment 1: Prediction Performance Comparison}

We evaluate the predictive performance of alphas generated by our method against baselines. To comprehensively assess the effectiveness of the generated alpha sets, we employ two representative machine learning models: LightGBM~\citep{ke2017lightgbm}, a popular gradient boosting framework known for its efficiency, and a 3-layer Multi-Layer Perceptron (MLP), which can capture complex non-linear relationships.
For each alpha generation method, we create alpha sets of three distinct sizes---10, 50, and 100---to serve as input features for these models. This allows for a thorough comparison of the mined alpha sets across varying sizes.
Both input alphas and target returns undergo cross-sectional rank normalization before training to mitigate outlier influence.
The predictive power of the alphas is evaluated using two standard metrics in quantitative finance: the Information Coefficient (IC) and the Rank Information Coefficient (RankIC).

\begin{table*}[t!]
    \centering
    \caption{Ablation study of our framework's components. 
    % We incrementally add multi-dimensional feedback criteria and the FSA mechanism.
    Best results are highlighted in \textbf{bold}. We ablate across five evaluation dimensions: Effectiveness (Eff.), Diversity (Div.), Turnover (Turn.), Stability (Stab.), and Overfitting Risk (O.R.). The checkmark (\checkmark) indicates a dimension is included in the search, while the cross (\texttimes) indicates it is not.}
    \label{tab:ablation_study}
    \resizebox{\textwidth}{!}{
    \begin{tabular}{@{}l ccccc cccc cccc@{}}
        \toprule
        \multirow{2.5}{*}{Search Strategy} & \multicolumn{5}{c}{Included Evaluation Dimensions} & \multicolumn{4}{c}{LightGBM} & \multicolumn{4}{c}{MLP} \\
        \cmidrule(lr){2-6} \cmidrule(lr){7-10} \cmidrule(lr){11-14}
        & Eff. & Div. & Turn. & Stab. & O.R. & IC & RankIC & AER & IR & IC & RankIC & AER & IR \\
        \midrule
        CoT & 
        \checkmark & \checkmark & \texttimes & \texttimes & \texttimes & 
        0.0434 & 0.0395 & 0.0707 & 0.7461 & 
        0.0421 & 0.0393 & 0.0922 & 0.9962 \\
        
        ToT & 
        \checkmark & \checkmark & \texttimes & \texttimes & \texttimes & 
        0.0459 & 0.0427 & 0.0868 & 0.9337 & 
        0.0452 & 0.0435 & 0.0945 & 1.0348 \\
        
        MCTS & 
        \checkmark & \texttimes & \texttimes & \texttimes & \texttimes & 
        0.0409 & 0.0374 & 0.0941 & 0.9775 & 
        0.0400 & 0.0376 & 0.0935 & 1.0010 \\
        
        MCTS & 
        \checkmark & \checkmark & \texttimes & \texttimes & \texttimes & 
        0.0501 & 0.0476 & 0.1003 & 1.0106 & 
        0.0486 & 0.0462 & 0.1023 & 1.0462 \\
        
        MCTS & 
        \checkmark & \checkmark & \checkmark & \texttimes & \texttimes & 
        0.0492 & 0.0457 & 0.1063 & 1.1062 & 
        0.0489 & 0.0462 & 0.1185 & 1.2556 \\
        
        MCTS & 
        \checkmark & \checkmark & \checkmark & \checkmark & \texttimes & 
        0.0495 & 0.0462 & 0.1030 & 1.0331 & 
        0.0491 & 0.0465 & 0.1093 & 1.1773 \\
        
        MCTS & 
        \checkmark & \checkmark & \checkmark & \checkmark & \checkmark & 
        0.0515 & 0.0479 & 0.1075 & 1.1121 & 
        0.0503 & 0.0478 & 0.1166 & 1.2127 \\
        
        MCTS+FSA & 
        \checkmark & \checkmark & \checkmark & \checkmark & \checkmark & 
        \textbf{0.0549} & \textbf{0.0512} & \textbf{0.1107} & \textbf{1.1792} & 
        \textbf{0.0522} & \textbf{0.0503} & \textbf{0.1234} & \textbf{1.2712} \\
        \bottomrule
    \end{tabular}
    }
\end{table*}

To assess the practical profitability of the mined alphas in simulated real-world stock market scenarios, we follow established evaluation methodologies \cite{yu2023generating} and conduct backtests. Specifically, we employ a top-$k$/drop-$n$ portfolio construction strategy, implemented on the Qlib platform \cite{yang2020qlib}. The detailed configuration of this strategy is described in Appendix~\ref{sec:backtesting_strategy}. The practical trading performance is evaluated using two key metrics: Annualized Excess Return (AER), which measures the strategy's profitability, and Information Ratio (IR), which quantifies its risk-adjusted performance.

The combination of these four metrics (IC, RankIC, AER, and IR) provides a comprehensive evaluation of the mined alphas.
As illustrated in Figure~\ref{fig:method_performance} (with detailed results presented in Appendix~\ref{sec:full_exp_result}), our framework consistently outperforms baselines across all metrics.
This demonstrates that the alphas mined by our framework possess superior predictive capabilities for future stock returns, which can be effectively translated into trading profitability. These findings are further corroborated by our experiments on the U.S. stock market, with detailed results presented in Appendix~\ref{sec:appendix_sp500}.

\begin{figure}[t]
  \vspace{-\intextsep}
  \centering
  \includegraphics[width=1.0\columnwidth]{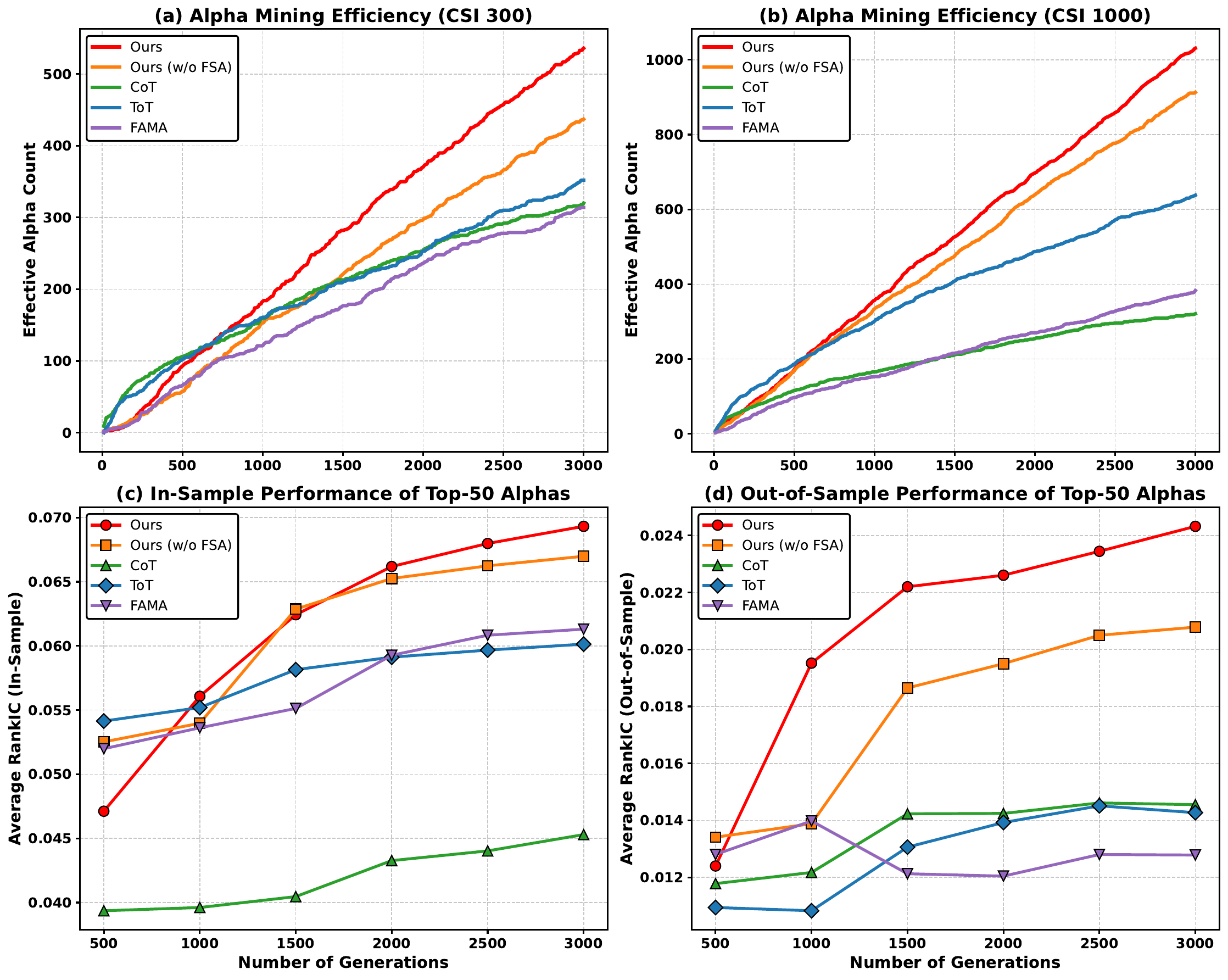}
  \caption{Analysis of search dynamics. (a, b) Alpha mining efficiency, measured as the count of effective alphas found versus total generated. (c, d) Average in-sample and out-of-sample RankIC of the top 50 alphas over generations.}
  \label{fig:search_dynamics}
  \vspace{-\intextsep}
\end{figure}

\subsection{Experiment 2: Ablation Study}
\label{sec:ablation_study}

We conduct ablation studies to evaluate three key components of our framework: MCTS, multi-dimensional feedback, and FSA.

Table~\ref{tab:ablation_study} presents the impact of these components on predictive performance.
When incorporating Effectiveness and Diversity as feedback, MCTS demonstrates superior predictive performance over CoT and ToT.
Performance progressively improves with the integration of additional feedback dimensions. Notably, while Turnover feedback slightly reduces IC and RankIC, it enhances practical trading metrics (AER, IR) by mitigating transaction costs.
The integration of FSA yields further improvements across all metrics for both LightGBM and MLP models.
These results underscore the individual and collective contributions of MCTS, multi-dimensional feedback, and FSA to our framework's efficacy.

Beyond predictive performance, we analyze the search dynamics in Figure~\ref{fig:search_dynamics}. Subplots (a) and (b) assess search efficiency by plotting the number of effective alphas mined against the total generated alphas. Our framework, even without FSA, demonstrates higher search efficiency than other LLM-based methods. FSA further amplifies this advantage. Furthermore, subplots (c) and (d) evaluate the quality and generalization of the top 50 alphas (selected by in-sample RankIC) over generations. The in-sample performance of our method's alphas improves as the search progresses (c). More importantly, this improvement translates to superior out-of-sample performance (d), indicating that our framework discovers more generalizable alphas compared to baselines.

\begin{figure}[t]
  \vspace{-\intextsep}
  \centering
  \includegraphics[width=1.0\columnwidth]{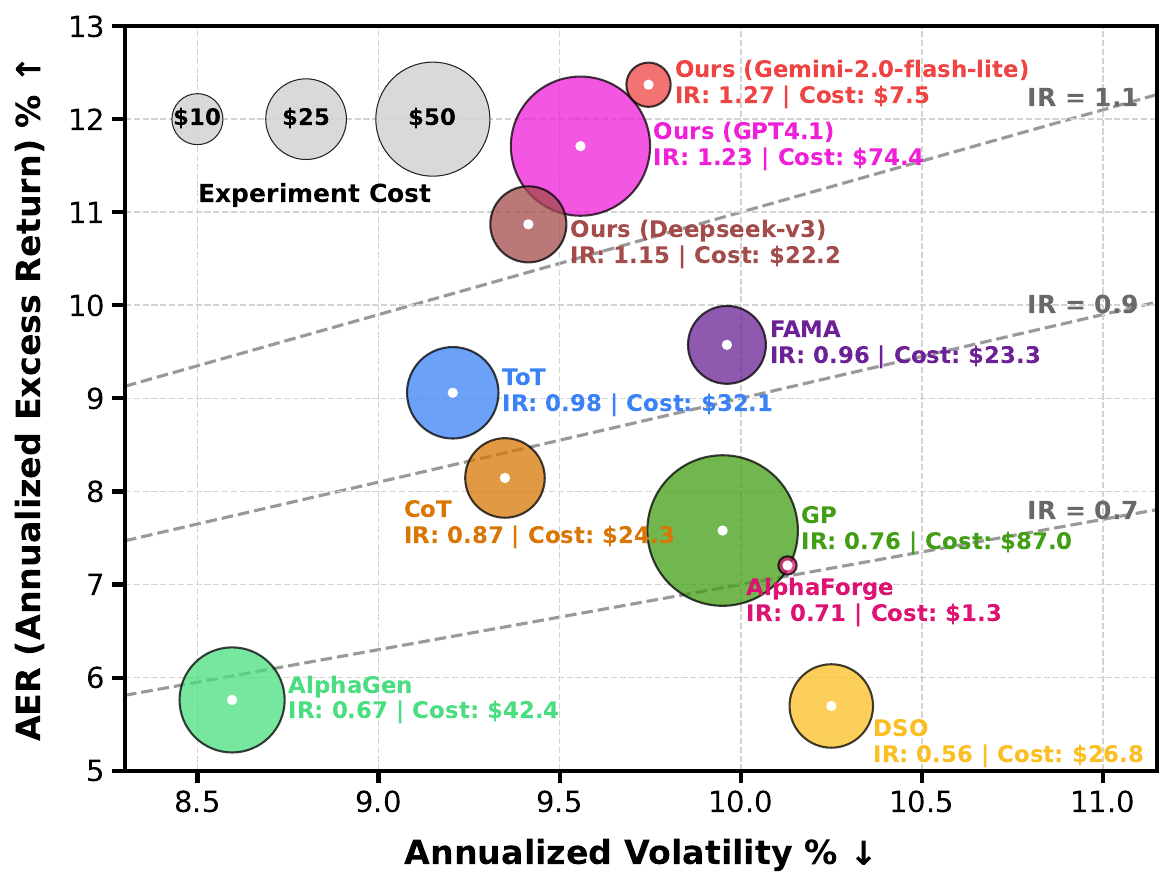}
  \caption{Cost-performance analysis across various methods. The plot shows Annualized Excess Return (AER) vs. Annualized Volatility (lower is better). Bubble size indicates the estimated single-run cost. Dashed lines represent constant Information Ratio (IR) levels.}
  \label{fig:cost_performance}
  \vspace{-\intextsep}
\end{figure}

To establish a more standardized and equitable basis for comparison, we present a cost-performance analysis in Figure~\ref{fig:cost_performance}. We estimate the cost of a single experimental run for each method by unifying its runtime and API usage into a monetary value based on public cloud computing prices (detailed in Appendix~\ref{sec:cost_estimation}). 
The analysis reveals that our framework achieves a favorable risk-return profile, with its variants occupying the highest Information Ratio (IR) contours. 
Notably, the overall cost and performance of our framework are primarily driven by the choice of the underlying LLM. 
For instance, employing a lightweight model like Gemini-2.0-flash-lite yields a high IR of 1.27 at a minimal cost of \$7.5. 
In contrast, using the more powerful GPT-4.1 results in a slightly lower IR of 1.23 at a substantially higher cost of \$74.4. 
This highlights that our framework offers the flexibility to select an appropriate LLM, enabling a desirable balance between performance and computational budget.

\subsection{Experiment 3: Interpretablitity of Alpha Formulas}

\begin{figure}[t]
  \vspace{-\intextsep}
  \centering
  \includegraphics[width=1.0\columnwidth]{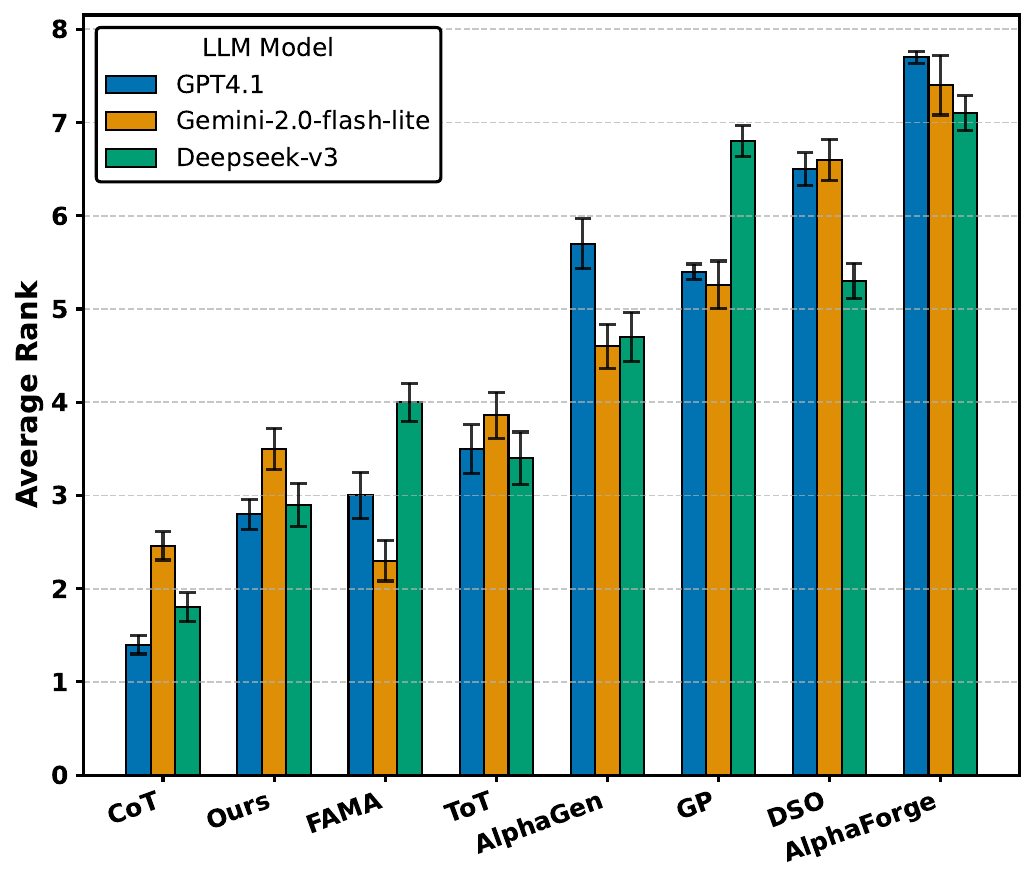}
  \caption{Interpretability Ranking Comparison, showing mean and standard deviation over 5 random seeds.}
  \label{fig:interp_comp}
  \vspace{-\intextsep}
\end{figure}

In this experiment, we evaluate the interpretability of alpha formulas mined by different methods. We define the interpretability of an alpha formula by its capacity to articulate a reasonable logic, a specific market phenomenon, or an investment strategy.
To quantify this, we randomly select one alpha formula per method and employ LLMs to rank their interpretability. We repeat this process 50 times and compute the average rank for each method. To mitigate potential biases from a single LLM, we aggregate rankings from three distinct LLMs. The results are presented in Figure~\ref{fig:interp_comp}. The findings indicate that formulas mined by our framework exhibit interpretability second only to those generated by the CoT method. 
This result is particularly noteworthy when compared to non-LLM baselines, whose formulas were consistently ranked as less interpretable.
This suggests that our approach achieves a compelling trade-off, delivering strong predictive performance while maintaining a high degree of interpretability.

Furthermore, acknowledging that LLMs might inherently favor formulas generated through LLM-driven processes, we provide illustrative examples of alpha formulas from our method and non-LLM-based approaches in Appendix~\ref{sec:interpretability_examples}.
While a formal human study is outside the scope of this work, a qualitative inspection of these examples reveals a clear difference: formulas from our method tend to have more discernible logic compared to the often opaque structures produced by non-LLM methods. 
% This superiority typically manifests as more discernible underlying logic and enhanced transparency when compared to formulas produced by non-LLM-based methods.

%% file: sections/Related_Work.tex
\section{Related Work}

Automated formulaic alpha mining has traditionally relied on genetic programming (GP) frameworks \cite{lin2019stock, zhang2020autoalpha}, with other notable approaches employing reinforcement learning \cite{yu2023generating} and deep learning-based generative models \cite{shi2024alphaforge}. More recently, LLMs have been introduced, with approaches like FAMA \cite{li2024can} and AlphaAgent \cite{tang2025alphaagent} leveraging them for direct alpha generation guided by in-context examples or heuristics. In contrast, our method frames alpha discovery as a formal reasoning task, uniquely employing MCTS to systematically explore the structured space of mathematical formulas. This approach is inspired by recent advancements in tree search-based reasoning, such as Tree of Thoughts (ToT) \cite{yao2024tree} and RethinkMCTS \cite{li2024rethinkmcts}, which enhance LLM planning by structuring generation as a search process. A further literature review is provided in Appendix A.

%% file: sections/Conclusion.tex
\section{Conclusion}

We introduce an LLM-Powered Monte Carlo Tree Search (MCTS) framework for formulaic alpha mining. This approach models alpha mining as a tree search, where an LLM iteratively generates and refines candidate formulas, critically guided by quantitative feedback from financial backtesting. To foster search efficiency and alpha effectiveness, we incorporate a Frequent Subtree Avoidance mechanism. Experimental results demonstrate that our framework mines alphas with superior predictive accuracy and trading performance, while also offering enhanced interpretability and search efficiency compared to existing methods. This work pioneers a promising direction for leveraging LLMs and MCTS to tackle the complex challenge of automated formulaic alpha mining in finance.

%% file: sections/Appendix.tex
\onecolumn
\section{Summary of Appendix}

This appendix provides extensive supplementary material to support the methodologies, experiments, and findings presented in the main paper. The contents are organized as follows:

\begin{itemize}
    \item \textbf{Section~\ref{sec:appendix_related_work}: Further Related Work.} A review of related work in automated formulaic alpha mining and tree search-based reasoning.
    \item \textbf{Section~\ref{sec:appendix_case_study}: Illustrative Case Study.} A step-by-step walkthrough of our framework's MCTS workflow.
    \item \textbf{Section~\ref{sec:appendix_method_details}: Method Details.} Elaboration on the generation and refinement of alpha formulas, the multi-dimensional evaluation metrics, and the dynamic search budget strategy.
    \item \textbf{Section~\ref{sec:appendix_operators}: List of Operators.} A comprehensive list of operators employed within the alpha formulas.
    \item \textbf{Section~\ref{sec:appendix_pseudo_code}: Pseudo-code.} The algorithmic pseudo-code of our framework.
    \item \textbf{Section~\ref{sec:appendix_exp_details}: Experimental Setup Details.} Details on the dataset, hyperparameter configurations, model specifications, backtesting strategy, and performance metrics.
    \item \textbf{Section~\ref{sec:appendix_additional_results}: Extended Experimental Results and Analyses.} Includes comparative analyses against other baselines, comparison results on the U.S. stock market, investigations into data leakage and LLM sensitivity, and qualitative analyses of alpha characteristics and interpretability.
    \item \textbf{Section~\ref{sec:full_exp_result}: Full Experiment Results.} Comprehensive results of the prediction performance comparison experiments.
    \item \textbf{Section~\ref{sec:limitation}: Limitations.} A discussion of the limitations of our framework.
    \item \textbf{Section~\ref{sec:appendix_prompts}: LLM Agent Prompts.} Details of key prompts used for alpha generation, refinement, and overfitting risk assessment.
\end{itemize}

\section{Further Related Work} \label{sec:appendix_related_work}

\subsection{Automated Formulaic Alpha Mining} \label{subsec:appendix_alpha_mining}

\paragraph{Traditional Methods} As mentioned in the main text, traditional alpha mining encompasses several paradigms. Genetic programming (GP) is a foundational approach. Early work like GPLearn \cite{lin2019stock} applies GP with a pre-defined set of time-series operators, while later enhancements like AutoAlpha \cite{zhang2020autoalpha} improve efficiency by initializing the search with effective, low-depth alphas. AlphaEvolve \cite{cui2021alphaevolve} further advances this line of work by evolving computation graphs instead of simple formula trees. Beyond GP, other paradigms include AlphaGen \cite{yu2023generating}, which uses reinforcement learning to optimize an entire alpha set, and AlphaForge \cite{shi2024alphaforge}, which designs a deep learning-based generative-predictive structure.

\paragraph{Other LLM-based Frameworks} Several other works have explored the application of LLMs in quantitative finance, though they fall outside the scope of our direct comparison. For instance, AlphaGPT \cite{wang2023alpha} focuses on a human-AI interaction paradigm, using prompt engineering to translate researcher insights into alphas. A framework by Kou et al. \cite{kou2024automate} mines alphas from multimodal financial data via a multi-agent system. Similarly, QuantAgent \cite{wang2024quantagent} introduces a two-loop system where an LLM refines strategies from a knowledge base updated through real-world testing. These methods are excluded from our comparative benchmarks due to their reliance on non-standard data (e.g., multimodal), interactive human feedback, or methodologies that are not sufficiently detailed or open-sourced for a fair and reproducible comparison.

\subsection{Tree Search-based Reasoning} \label{subsec:appendix_tree_search}

As mentioned in the main text, tree search methods are increasingly used to maximize the exploration capabilities of LLMs, allowing for different levels of search and planning \cite{zhang2023planning,hu2024uncertainty}. They are now widely applied in LLM-based agents and reasoning tasks \cite{wang2024math,wang2024q}. LATS \cite{zhou2023language}, for example, extends this concept by viewing the LLM as a general agent that conducts exploration at both the reasoning and action levels, which is conceptually related to our approach of building a formula through a sequence of operator selections.

\section{Illustrative Case Study of MCTS Workflow}
\label{sec:appendix_case_study}

To elucidate the workflow of our proposed MCTS framework for alpha factor mining, this section provides a detailed step-by-step example.

\begin{figure*}[htbp]
    \centering
    \includegraphics[width=1.0\textwidth]{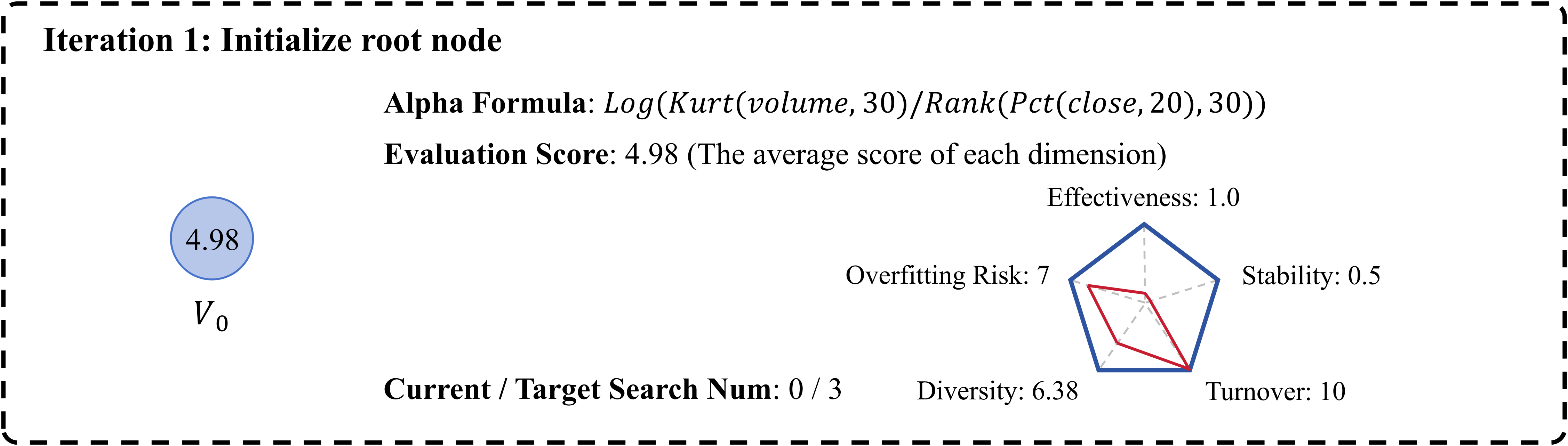}
    \caption{Case Study: MCTS root node generation.}
    \label{fig:mcts_root}
\end{figure*}

Initially, as depicted in Figure~\ref{fig:mcts_root}, an initial alpha formula is generated by the LLM. After undergoing a multi-dimensional evaluation, this formula becomes the root node ($V_0$) of the MCTS search tree. For this illustrative case, the target search count (initial search budget) is initialized to 3. This target is dynamically incremented by 1 (budget increment) each time a newly generated node achieves a new high score, encouraging deeper exploration of promising search trees.

\begin{figure*}[htbp]
    \centering
    \includegraphics[width=1.0\textwidth]{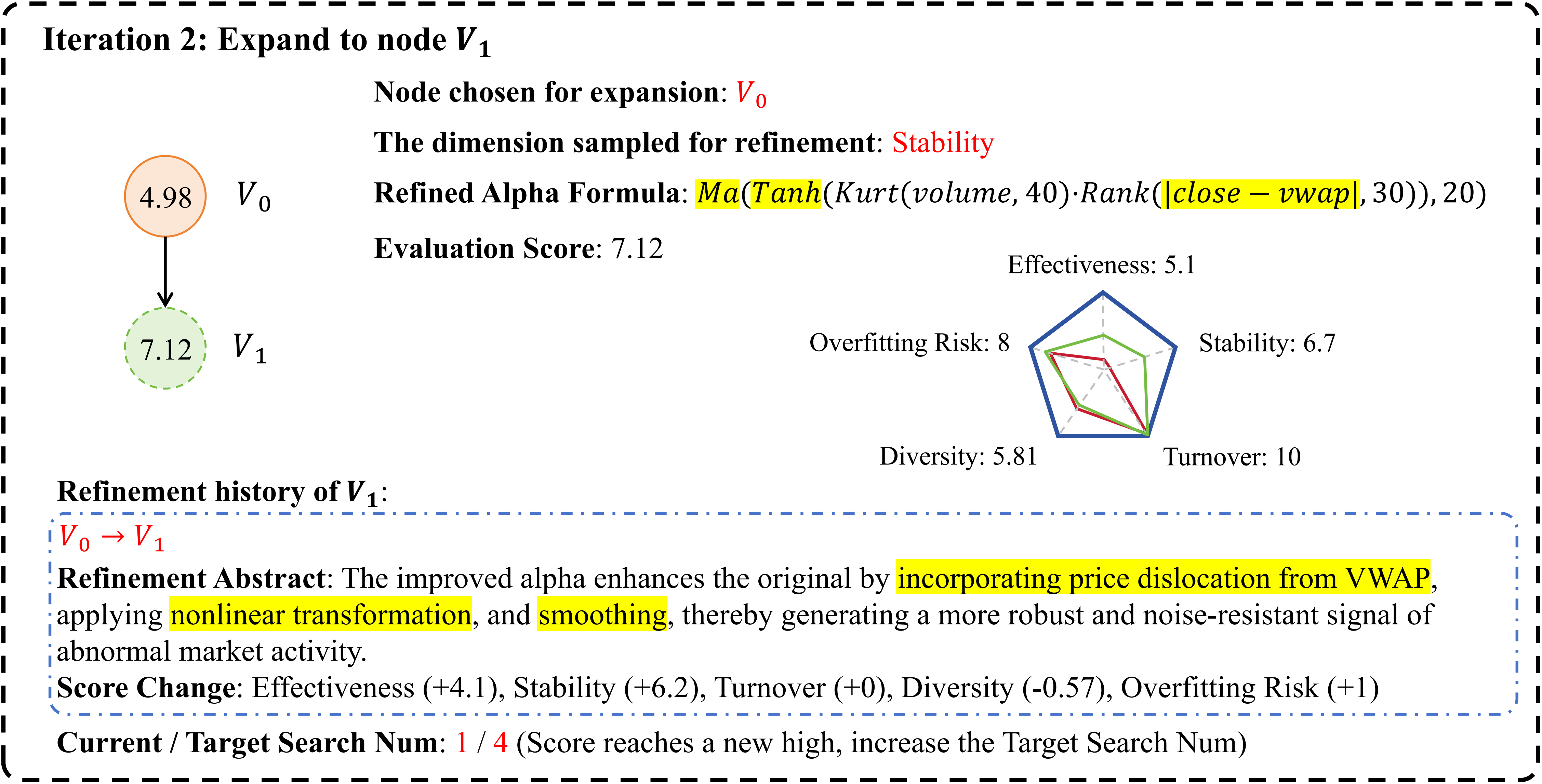}
    \caption{Case Study: expansion to node $V_1$.}
    \label{fig:mcts_v1}
\end{figure*}

Following the construction of the root node, the MCTS process proceeds with node expansion. As shown in Figure~\ref{fig:mcts_v1}, the root node $V_0$ is expanded to generate its first child node, $V_1$. The dimension for refinement is selected by sampling according to Equation~\eqref{eq:refinement_sampling_criterion} (in this instance, the \textit{Stability} dimension is chosen). Subsequently, the LLM generates targeted refinement suggestions. Based on these suggestions, a refined alpha formula is produced, with the modifications highlighted within the formula representation. This refined alpha formula is then subjected to the same multi-dimensional evaluation process, yielding the scored node $V_1$. The refinement history for $V_1$ is updated to include a summary of this specific refinement step and the corresponding change in evaluation scores. Finally, both the target search count (if $V_1$'s score is a new high) and the current search count are updated.

\begin{figure*}[htbp]
    \centering
    \includegraphics[width=1.0\textwidth]{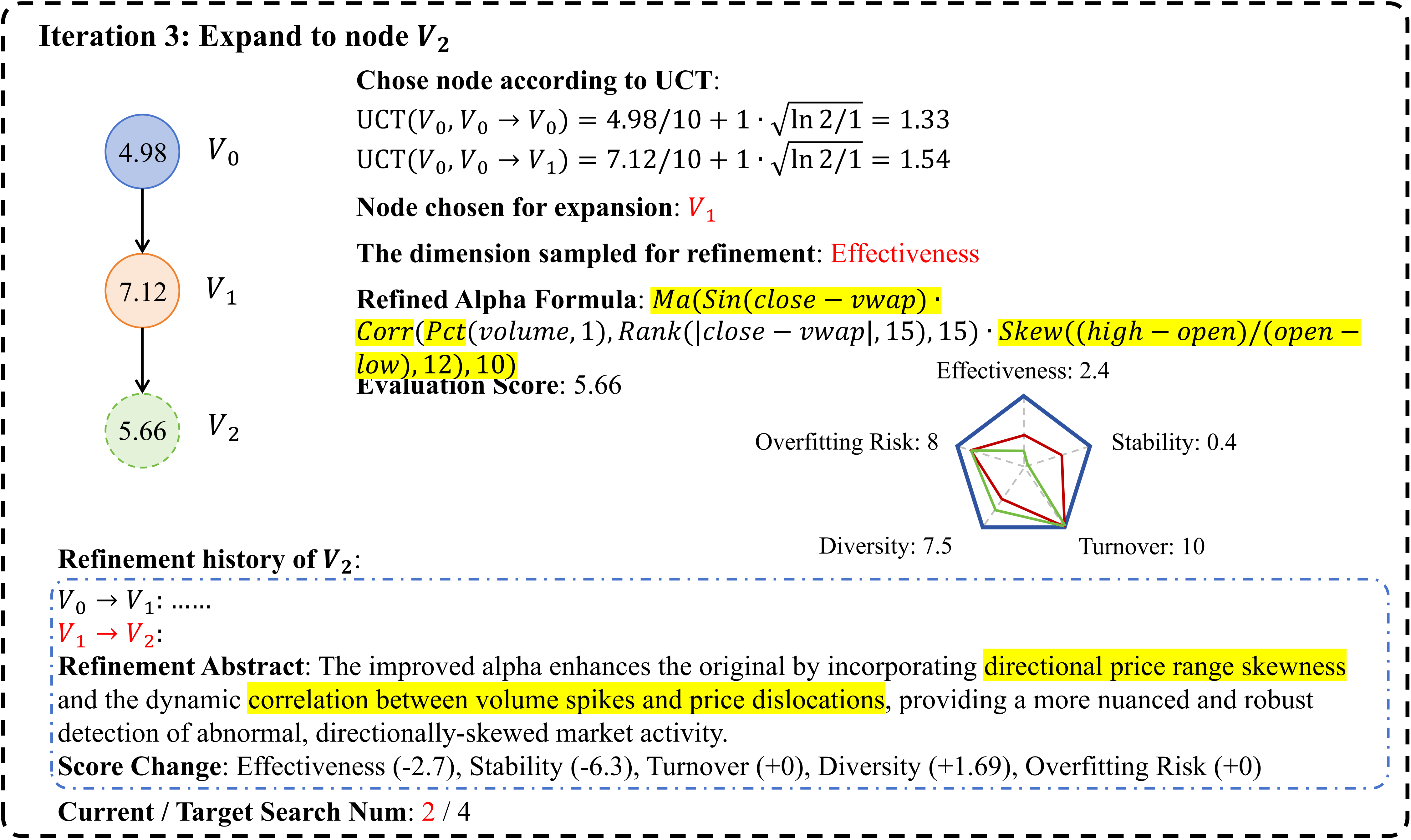}
    \caption{Case Study: expansion to node $V_2$.}
    \label{fig:mcts_v2}
\end{figure*}

The expansion process continues, as illustrated in Figure~\ref{fig:mcts_v2}. Based on the UCT criterion, node $V_1$ is selected for the next expansion. Similar to the previous step, a dimension for refinement is sampled. The LLM then generates an improved alpha formula based on this dimension, which is subsequently evaluated to create node $V_2$. Notably, the refinement history of $V_2$ is cumulative, encapsulating all refinement steps and score evolutions from the root node $V_0$ through $V_1$ to $V_2$. The current search count is then incremented.

\begin{figure*}[htbp]
    \centering
    \includegraphics[width=1.0\textwidth]{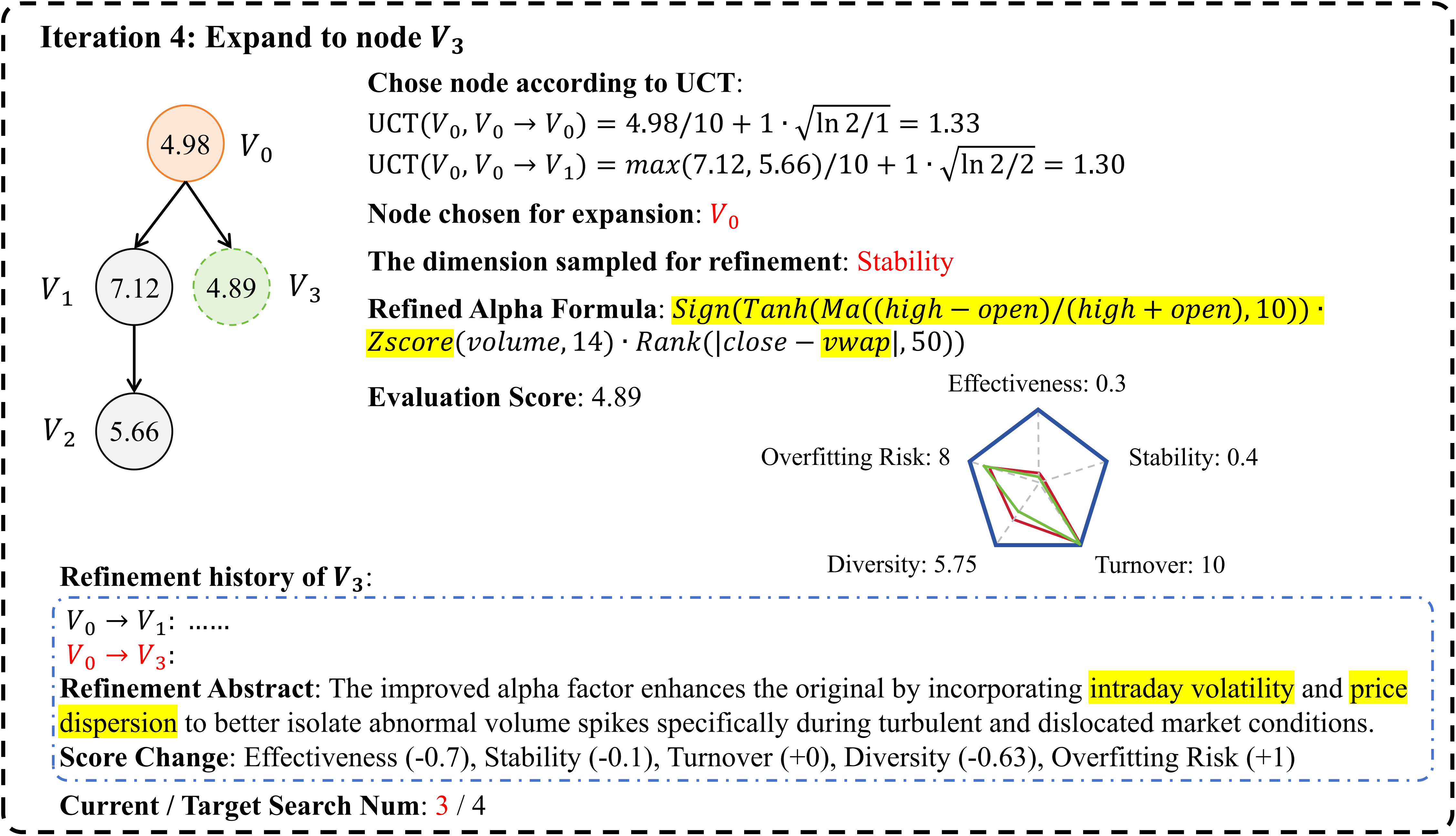}
    \caption{Case Study: expansion to node $V_3$.}
    \label{fig:mcts_v3}
\end{figure*}

\begin{figure*}[htbp]
    \centering
    \includegraphics[width=1.0\textwidth]{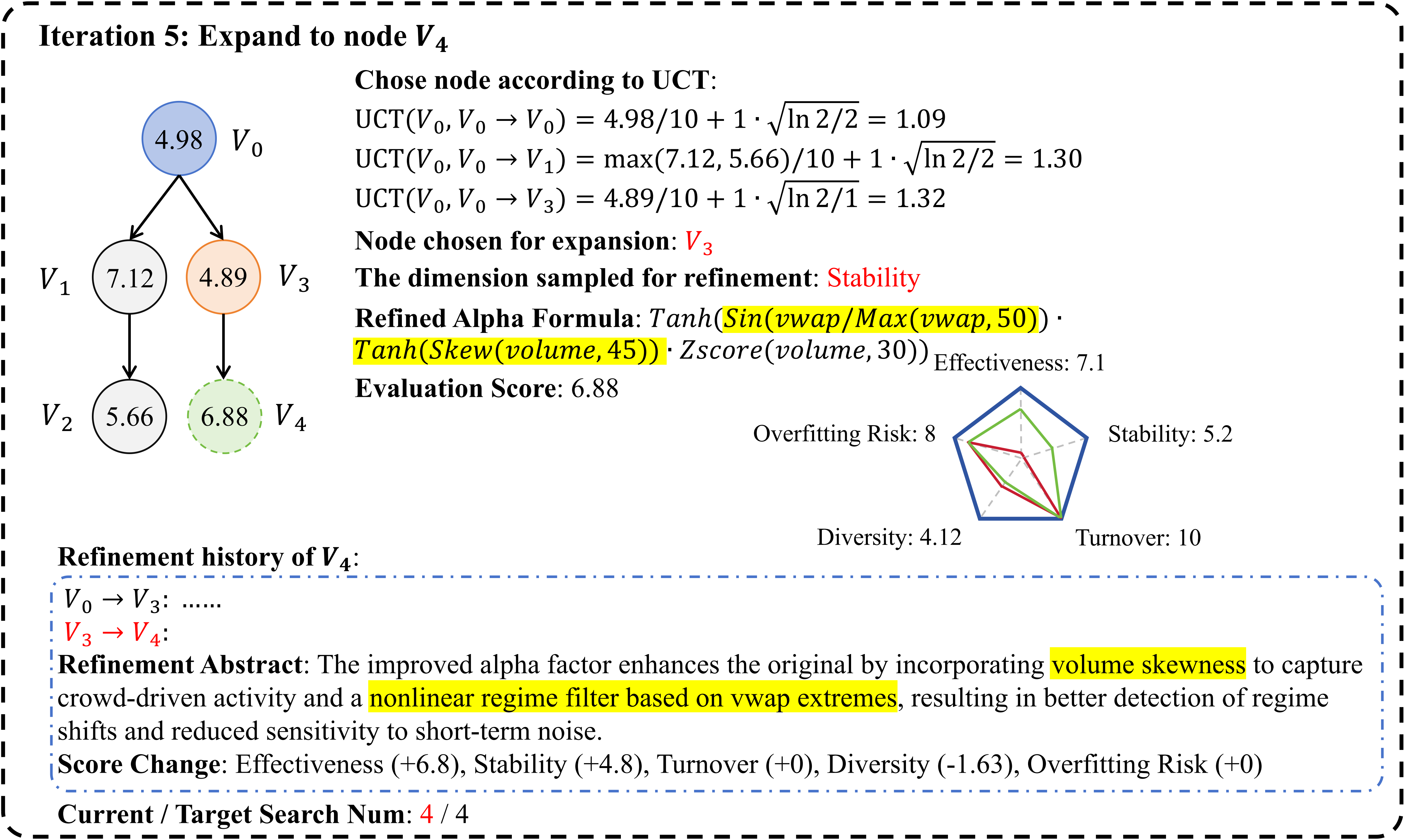}
    \caption{Case Study: expansion to node $V_4$ and search termination.}
    \label{fig:mcts_v4}
\end{figure*}

Figures~\ref{fig:mcts_v3} and \ref{fig:mcts_v4} showcase the outcomes of the two subsequent expansion steps, leading to the generation of nodes $V_3$ and $V_4$, respectively. Upon the expansion that creates node $V_4$, the current search count reaches the predetermined target. At this point, the expansion phase for the current MCTS iteration concludes. All five generated nodes ($V_0,...,V_4$) are then systematically reviewed. Nodes that meet predefined quality and performance criteria are added to the effective alpha repository.

This sequence completes one full iteration of the MCTS-driven alpha mining process. Our framework is designed to iteratively repeat this entire cycle, enabling the efficient and systematic mining of effective alphas.

\section{Method Details} \label{sec:appendix_method_details}

In this section, we elaborate on specific details of our framework, focusing on the generation process, evaluation metrics, and the search budget allocation strategy.

\subsection{Alpha Formula Generation} We decompose the generation of alpha formulas into a two-step process: first, an \textit{alpha portrait} is generated, and subsequently, the corresponding alpha formula is derived from this portrait. An alpha portrait is a textual representation that includes the alpha's name, a concise description of its underlying investment logic, and its formula expressed in pseudo-code. This two-step approach decouples the conceptual design of an alpha from its concrete construction, thereby reducing the instruction-following complexity for the LLM and enhancing the quality of the generated formulas.

When generating alpha formulas, the LLM is instructed to use symbolic parameters (e.g., the lookback window for a moving average operator) rather than fixed numerical values. Concurrently, the LLM proposes several candidate sets of parameter values (three sets in our experiments). We then backtest the alpha derived from each parameter set and select the configuration yielding the best performance. This strategy facilitates a more efficient exploration and utilization of each generated formula structure.

\subsection{Refinement Suggestion Generation} \label{sec:refinement_suggestion_generation}

During each refinement iteration, once a target dimension for improvement is selected as per Equation~\ref{eq:refinement_sampling_criterion}, we leverage the LLM to generate targeted refinement suggestions. We employ few-shot learning, using alphas from the effective alpha repository $\mathcal{F}_{zoo}$ as exemplars. The exemplar selection strategy is tailored for each evaluation dimension to provide the most relevant guidance:

\paragraph{Effectiveness and Stability} To select diverse yet potent exemplars, we first filter the effective alpha repository by removing alphas with the highest correlation to the current alpha (top $\eta$\%, where $\eta=50$ in our experiments). This creates a candidate pool of more structurally distinct exemplars. From this filtered subset, we select the top-$k$ alphas (e.g., $k=3$) exhibiting the highest scores in Effectiveness or Stability to serve as few-shot examples. This two-stage process ensures diversity by preventing highly similar alphas from dominating the exemplar set and encouraging more innovative refinements.

\paragraph{Diversity} To encourage the exploration of novel alpha structures, we select the top-$k$ alphas from the effective alpha repository $\mathcal{F}_{zoo}$ that exhibit the lowest correlation with the current alpha as few-shot exemplars.

\paragraph{Turnover and Overfitting Risk} For these dimensions, we adopt a zero-shot approach, directly prompting the LLM to generate refinement suggestions without explicit exemplars, relying on the model's intrinsic understanding of these concepts.

Following the generation of a textual suggestion, the LLM produces a revised alpha formula. We then perform a syntax validation on this formula. If it is syntactically incorrect, the LLM receives the error as feedback and is prompted to correct it iteratively until the formula passes validation.

\subsection{Multi-Dimensional Evaluation Metrics} \label{sec:appendix_evaluation_metrics}

As introduced in Section~\ref{sec:evaluation}, we evaluate each candidate alpha $f$ across a set of $q$ dimensions to provide granular feedback for refinement. The score for each dimension, $e_i$, is typically derived from its percentile rank against the repository $\mathcal{F}_{zoo}$ as defined in Equation~\ref{eq:dim_score}. The primary dimensions used in our framework are detailed as follows:

\begin{itemize}
    \item Effectiveness: This dimension measures the alpha's core predictive power.
    \item Stability: This dimension assesses the consistency of an alpha's predictive performance over time. An alpha that performs well only in specific market regimes is less reliable.
    \item Turnover: This dimension evaluates the trading cost associated with an alpha. It is measured by the average daily change in the alpha's portfolio holdings. A high turnover implies frequent rebalancing, which can incur significant transaction costs and erode profits. The goal is to maintain turnover within a desirable, low range.
    \item Diversity: This dimension quantifies the novelty an alpha contributes to the existing repository $\mathcal{F}{zoo}$. A valuable alpha should provide predictive signals that are not redundant with those already discovered.
    \item Overfitting Risk: As described in Section~\ref{sec:evaluation}, this is a qualitative assessment performed by an LLM. The model analyzes the formula's complexity, the number of parameters, and its refinement history to identify signs of being overly tailored to the training data (i.e., ``p-hacking''), which would impair its generalization performance.
\end{itemize}

\subsection{Dynamic Search Budget Allocation} To balance the breadth and depth of the MCTS search, we employ a dynamic budget allocation strategy. This approach directs more computational resources towards the most promising search paths without prematurely abandoning exploration. The mechanism is governed by two parameters: an initial search budget $B$ and a budget increment $b$.

The search process for any given MCTS tree begins with an initial budget of $B$ expansion steps. During the search, we monitor the score of each newly generated alpha. If a new alpha $f_{new}$ achieves a score $S(f_{new})$ that surpasses the highest score previously seen within that specific MCTS tree, it constitutes a ``breakthrough''. Upon each such breakthrough, the remaining search budget for the current tree is increased by the increment $b$. The search for the current tree terminates when its allocated budget is exhausted.

This strategy ensures that promising avenues, identified by the discovery of high-quality alphas, are explored more thoroughly, while preventing excessive resource expenditure on less fruitful branches of the search space $\mathcal{A}$. It dynamically adapts the search effort based on real-time performance feedback, leading to a more efficient and effective alpha mining process.

\section{List of Operators}

In this paper, the operators we used can be categorized into two classes: unary operators and binary operators. All operators perform temporal operations, utilizing data from either the current trading day alone or including data from a preceding period. We list the details of all the operators in Table \ref{tab:operators}

\begin{table}[htbp]
    \centering
    
    \resizebox{\textwidth}{!}{
    \begin{tabular}{l|c|l}
        \toprule
        \textbf{Operator} & \textbf{Type} & \textbf{Description} \\
        \midrule
        $-x,|x|,x^2,1/x$ &  \multirow{15}{*}{\makecell{Unary}} & The opposite/absolute/square/inverse value of $x$. \\
        $\mathrm{Sign}(x)$ &   & The sign of $x$. \\
        $\mathrm{Sin}(x),\mathrm{Cos}(x),\mathrm{Tanh}(x)$ &  & Sine, cosine, and hyperbolic tangent functions.\\
        $\mathrm{Log}(x)$ &  & The natural logarithm of $x$. \\
        $\mathrm{Delay}(x,t)$ &  & The value of expression $x$ at $t$ trading days prior. \\
        $\mathrm{Diff}(x,t)$ &  & The difference between x and the value prior to day t, $x-\mathrm{Delay}(x,t)$. \\
        $\mathrm{Pct}(x,t)$ &  & The rate of change of x relative to its value t days prior. \\
        $\mathrm{Ma}(x,t),\mathrm{Med}(x,t),\mathrm{Sum}(x,t)$ &  & The mean/median/sum value of $x$ over the past $t$ days. \\
        $\mathrm{Std}(x,t)$ &  & The standard deviation value of $x$ over the past $t$ days. \\
        $\mathrm{Max}(x,t),\mathrm{Min}(x,t)$ &  & The maximum/minimum value of $x$ over the past $t$ days.\\
        $\mathrm{Rank}(x,t)$ &  & The ranking of $x$ relative to its values over the past $t$ days. \\
        $\mathrm{Skew}(x,t),\mathrm{Kurt}(x,t)$ &  & The skewness/kurtosis value of $x$ over the past $t$ days. \\
        $\mathrm{Vari}(x,t)$ &  & The variation value of $x$ over the past $t$ days, $\mathrm{Std}(x,t)/\mathrm{Ma}(x,t)$. \\
        $\mathrm{Autocorr}(x,t,n)$ &  & The autocorrelation coefficient of $x$ with a lag of $n$ over the past $t$ days. \\
        $\mathrm{Zscore}(x,t)$ &  & \makecell[l]{The z-score normalization of $x$ based on the mean and standard deviation\\ of its values over the past $t$ days.} \\
        \midrule
        $x+y,x-y,x\cdot y,x/y$ & \multirow{4}{*}{\makecell{Binary}} & Arithmetic operators. \\
        $\mathrm{Greater}(x,y),\mathrm{Less}(x,y)$ &  & Whether the first value is larger/smaller than the second value. \\
        $\mathrm{Cov}(x,y,t)$ &  & The covariance between time series $x$ and $y$ over the past $t$ days. \\
        $\mathrm{Corr}(x,y,t)$ &  & \makecell[l]{The Pearson's correlation coefficient between time series $x$ and $y$ \\ over the past $t$ days.} \\
        \bottomrule
    \end{tabular}}
    \caption{All the operators used in the experiments.}
    \label{tab:operators}
\end{table}

\label{sec:appendix_operators}

\section{Pseudo-Code}
\label{sec:appendix_pseudo_code}

We provide the pseudo-code for our LLM-guided MCTS framework in Algorithm~\ref{alg:llm_mcts_algo}.

\begin{algorithm}[htbp]
\caption{LLM-Guided MCTS Framework for Automated Alpha Mining}
\label{alg:llm_mcts_algo}
\small

\KwIn{$f_{\text{seed}}$ (initial alpha), $LLM$ (language model), $c$ (exploration const.), $T$ (temperature),
      $e_{\text{max}}$ (max score/dim), $B$ (initial search budget), $b$ (budget increment), $\theta_{\text{eff}}$ (effectiveness threshold),
      $\mathcal{G}_{\text{forbidden}}$ (initial forbidden structures)}
\KwOut{Repository $\mathcal{F}_{zoo}$ of effective alphas}

\BlankLine

\algphase{Initialization}
$\mathcal{F}_{zoo} \gets \emptyset$\;
$s_0 \gets \text{CreateRootNode}(f_{\text{seed}})$\;
$\boldsymbol{E}_{s_0} \gets \text{MultiDimEvaluate}(f_{\text{seed}}, \mathcal{F}_{zoo})$\;
$N_{s_0} \gets 1$; $S_{s_0} \gets \text{AggregateScore}(\boldsymbol{E}_{s_0})$ \Comment{Store the alpha's intrinsic score $S(f_{s_0})$}
Tree $\mathcal{T} \gets \{s_0\}$\;
$S_{\text{max}} \gets S_{s_0}$ \Comment{Track highest alpha score $S(f)$ found in the tree}\;

\For{$iter \gets 1$ \KwTo $B$}{
    \algphase{Selection Phase}
    $s_{\text{selected}}, P_{\text{path}} \gets \text{SelectNodeViaUCT}(\mathcal{T}, s_0, c)$ \Comment{Select leaf/internal node for expansion. $P_{\text{path}}$ is a list of $(s,a)$ pairs.}
    \BlankLine

    \algphase{Expansion Phase}
    \subphase{1. Prioritize refinement dimension}
    $\boldsymbol{E}_{s} \gets s_{\text{selected}}.\boldsymbol{E}$\;
    $P(i|s) \gets \text{Softmax}((e_{\text{max}}\mathbf{1}_q - \boldsymbol{E}_{s})/T)$\;
    $i^* \gets \text{SampleDimension}(P(i|s))$\;
    \BlankLine

    \subphase{2. LLM generates refinement and new alpha formula}
    $\text{context} \gets \text{GetNodeRefinementContext}(s_{\text{selected}}, \mathcal{T})$ \Comment{Parent, children, siblings history}
    $\text{examples} \gets \text{SampleEffectiveAlphas}(\mathcal{F}_{zoo})$ \Comment{Few-shot examples}
    $d_{s, i^*}, f_{new} \gets LLM.\text{GenerateRefinedAlpha}(s_{\text{selected}}.f, i^*, \text{context}, \text{examples}, \mathcal{G}_{\text{forbidden}})$\;
    \BlankLine

    \subphase{3. Validate and revise formula iteratively}
    \While{$\neg \text{IsValid}(f_{new})$}{
        $\text{feedback} \gets \text{GetInvalidityReason}(f_{new})$\;
        $d_{s, i^*}, f_{new} \gets LLM.\text{CorrectAlphaFormula}(d_{s, i^*}, \text{feedback}, \text{context}, \text{examples}, \mathcal{G}_{\text{forbidden}})$\;
    }
    \BlankLine

    \algphase{Evaluation of New Alpha}
    $\boldsymbol{E}_{new} \gets \text{MultiDimEvaluate}(f_{new}, \mathcal{F}_{zoo})$ \Comment{As in Section~\ref{sec:evaluation}, using Eq.~\ref{eq:relative_rank} and LLM for overfitting}
    $S(f_{new}) \gets \text{AggregateScore}(\boldsymbol{E}_{new})$ \Comment{Calculate overall score using Eq.~\ref{eq:overall_score}}\;
    \BlankLine

    \subphase{Create new node and add to tree}
    $s_{new} \gets \text{CreateNode}(f_{new}, \boldsymbol{E}_{new}, S(f_{new}))$\;
    $\text{AddChildNode}(\mathcal{T}, s_{\text{selected}}, s_{new})$ \Comment{The action leading to $s_{new}$ is implicitly defined}\;
    \BlankLine

    \algphase{Backpropagation Phase}
    \For{$(s_k, a_k)$ in $P_{\text{path}}$}{
        $N_{s_k} \gets N_{s_k} + 1$\;
        $Q(s_k, a_k) \gets \max(Q(s_k, a_k), S(f_{new}))$ \Comment{Update Q-value as per Eq.~\ref{eq:q_update}}
    }
    $N_{s_{new}} \gets N_{s_{new}} + 1$\;
    \BlankLine

    \algphase{Repository and System Updates}
    \If{$\text{GetEffectivenessScore}(\boldsymbol{E}_{new}) \ge \theta_{\text{eff}}$}{
        $\mathcal{F}_{zoo} \gets \mathcal{F}_{zoo} \cup \{f_{new}\}$\;
        $\mathcal{G}_{\text{forbidden}} \gets \text{UpdateForbiddenStructures}(\mathcal{F}_{zoo})$ \Comment{Update based on FSA logic}
    }
    \If{$S(f_{new}) > S_{\text{max}}$}{
        $B \gets B + b$ \Comment{Increase total search budget}
        $S_{\text{max}} \gets S(f_{new})$ \Comment{Update overall max score}
    }
}
\Return $\mathcal{F}_{zoo}$\;
\end{algorithm}

\section{Experimental Setup Details}
\label{sec:appendix_exp_details}

\subsection{Dataset}

All empirical data in this study are obtained through the Qlib platform~\cite{yang2020qlib}, an open-source framework for quantitative financial research. We utilize the daily-frequency data for the Chinese A-shares market as provided by Qlib.

To ensure maximum reproducibility and transparency, we deliberately refrain from applying any additional data filtering, preprocessing, or adjustments. The data is thus used in its original form as sourced from the platform. Furthermore, to maintain a consistent experimental environment, all backtesting simulations are implemented and executed within the Qlib framework. This leverages its integrated backtesting engine and ensures a standardized evaluation process for all experiments.

\subsection{Hyperparameter Configurations}

\paragraph{Temperature Parameter of LLMs} When generating alpha portraits and alpha formulas, the temperature is set to 1.0; when correcting illegal alpha formulas, the temperature is adjusted to 0.8; and when scoring the overfitting risk of alpha, the temperature is set to 0.1.

\paragraph{MCTS} We set the exploration weight $c$ in the UCT criterion to 1. For each search tree, the initial search budget is 3, and whenever a node achieves a higher score than previously, the search budget is increased by 1.

\paragraph{Effective Alpha Check} Upon the completion of a search tree's expansion, we examine the effectiveness of the alpha formulas corresponding to all nodes within the tree. Those alphas that pass the effectiveness check are added to the effective alpha repository. The specific criteria for determining effectiveness are as follows:

\begin{itemize}
    \item Basic Criteria: $\text{RankIC}\ge 0.015$, $\text{RankIR} \ge 0.3$, $R_{\mathrm{RankIC}} \le 0.95$, $R_{\mathrm{RankIR}} \le 0.95$.
    \item Turnover Criteria: $\text{Daily Turnover} \le 1.6$.
    \item Diversity Criteria: The maximum correlation with the alpha within the effective alpha repository is less than 0.8.
\end{itemize}

After the mining process is completed, we select the top $k$ alphas with the highest RankIR from the effective alpha repository to form the final alpha set. In our experiments, $k$ is set to 10, 50, and 100.

\paragraph{Evaluation Score} Here, we present the backtesting metrics utilized to calculate scores for evaluation dimensions: Effectiveness: RankIC, Stability: RankIR, Turnover: Daily turnover rate, Diversity: Maximum correlation with alphas in the effective alpha repository.

\paragraph{Other Settings} When generating alpha refinement suggestions, the number of few-shot examples is set to 1; in the Frequent Subtree Avoidance method, the number of frequent subtrees to be avoided is set to 3. When calculating the selection probability for each evaluation dimension, the temperature parameter is set to $T=1$. Furthermore, during the generation of refinement suggestions for the Effectiveness and Stability dimensions, the correlation filtering ratio is maintained at $\eta=50\%$.

\subsection{Model Settings}
To ensure a fair comparison when evaluating alpha pools derived from different methods, we utilize fixed hyperparameters and consistent training strategies for all models.

For the LightGBM model, we configure it with 32 leaves per tree and a total of 200 estimators. The maximum depth of each tree is limited to 8, and the learning rate is set to 0.05. Both L1 and L2 regularization coefficients are fixed at 0.1 to control model complexity. In line with a simple hold-out approach, the model is trained on the entire training dataset (2011/01/01--2020/12/31) for the predetermined number of boosting rounds, without using a separate validation set for early stopping. The fully trained model is then evaluated on the final test set.

Similarly, we train the Multi-Layer Perceptron (MLP) models with a consistent configuration. The MLP architecture consists of three hidden layers with 256, 128, and 64 units, respectively. A dropout rate of 0.3 is applied after each hidden layer to mitigate overfitting. We employ the Adam optimizer with a learning rate of 0.001 and a batch size of 1024, using Mean Squared Error (MSE) as the loss function. To prevent overfitting, early stopping with a patience of 5 epochs is implemented. For this purpose, we partition the data, designating the final year of the training period (2020/01/01--2020/12/31) as the validation set. The model is therefore trained on data from 2011 through 2019, with its performance on the validation set monitored to determine the stopping point.

\subsection{Backtesting Strategy}
\label{sec:backtesting_strategy}

The top-$k$/drop-$n$ portfolio construction strategy employed in our backtests operates as follows. On each trading day, an equal-weight portfolio is formed by selecting the top $k$ stocks based on the predictive signals from the trained models. We set $k$ to represent the top 10\% of the respective stock pool (e.g., $k=30$ for the CSI300 index and $k=100$ for the CSI1000 index). To manage turnover and limit trading costs, a maximum of $n$ stocks are bought or sold daily. The value of $n$ is determined by $n = k/w$, where $w$ is the prediction horizon in days (e.g., for a 10-day return prediction on CSI300, $n=30/10=3$). This strategy ensures that the theoretical complete turnover period of the portfolio aligns with the prediction horizon. A conservative transaction cost of $0.15\%$ per trade is incorporated to ensure a realistic performance assessment.

\subsection{Environment}
\label{sec:exp_environment}

All the experiments are conducted with following settings:

\begin{itemize}
    \item CPU: AMD EPYC 7642 48-Core Processor
    \item GPU: NVIDIA GeForce RTX 3080Ti
    \item Operating system: Ubuntu 20.04.3 LTS
    \item Software versions: Python 3.8.5; Numpy 1.24.4; Pandas 1.5.2; Pytorch 2.2.2; Openai 1.57.4
\end{itemize}

\subsection{Predictive Performance Evaluation Metrics}

To evaluate the predictive performance of the model (trained on the mined alpha set) in forecasting stock returns, we employ several standard metrics: the Information Coefficient (IC), Rank Information Coefficient (RankIC), Annualized Excess Return (AER), and Information Ratio (IR). Let $f_{i,t}$ denote the predicted returns for asset $i$ at time $t$ (out of $N_t$ assets in the universe at time $t$), and $r_{i,t+1}$ be its realized total return over the subsequent period (e.g., from $t$ to $t+1$). The evaluation spans a total of $T$ time periods.

\paragraph{Information Coefficient (IC)}
The IC measures the linear correlation between predicted returns and subsequent total returns. It is calculated for each cross-section at time $t$ ($\text{IC}_t$) and then averaged over all $T$ periods:
\begin{align}
    \text{IC}_t &= \frac{\sum_{i=1}^{N_t} (f_{i,t} - \bar{f}_t)(r_{i,t+1} - \bar{r}_{t+1})}{\sqrt{\sum_{i=1}^{N_t} (f_{i,t} - \bar{f}_t)^2} \sqrt{\sum_{i=1}^{N_t} (r_{i,t+1} - \bar{r}_{t+1})^2}} \\
    \text{IC} &= \frac{1}{T} \sum_{t=1}^{T} \text{IC}_t
\end{align}
where $\bar{f}_t$ and $\bar{r}_{t+1}$ are the cross-sectional means of predicted signals and realized total returns, respectively. A higher IC indicates better predictive power.

\paragraph{Rank Information Coefficient (RankIC)}
The RankIC measures the monotonic relationship (Spearman's rank correlation) between predicted returns and subsequent total returns. It is less sensitive to outliers than the Pearson correlation-based IC. Similar to IC, it is computed cross-sectionally and then averaged:
\begin{align}
    \text{RankIC}_t &= \text{Corr}(\text{rank}(f_{1,t}, \dots, f_{N_t,t}), \text{rank}(r_{1,t+1}, \dots, r_{N_t,t+1})) \\
    \text{RankIC} &= \frac{1}{T} \sum_{t=1}^{T} \text{RankIC}_t
\end{align}
where $\text{rank}(\cdot)$ denotes the operation that assigns ranks to the elements in the vector, and $\text{Corr}(\cdot, \cdot)$ is the Pearson correlation coefficient applied to these ranks.

\begin{table}[t!]
\centering

\setlength{\tabcolsep}{4pt}
\begin{tabular}{@{}c|c| l |r r r r |r r r r@{}}
\toprule
\multirow{2}{*}{Instruments} & \multirow{2}{*}{\textbf{$\Delta T$}} & \multirow{2}{*}{Alpha Set} & \multicolumn{4}{c|}{LightGBM} & \multicolumn{4}{c}{MLP} \\
\cmidrule(lr){4-7} \cmidrule(lr){8-11}
& & & \multicolumn{1}{c}{IC} & \multicolumn{1}{c}{RankIC} & \multicolumn{1}{c}{AER} & \multicolumn{1}{c|}{IR} & \multicolumn{1}{c}{IC} & \multicolumn{1}{c}{RankIC} & \multicolumn{1}{c}{AER} & \multicolumn{1}{c}{IR} \\
\midrule
\multirow{6}{*}{CSI300}
& \multirow{3}{*}{10} & Alpha158  & 0.0386 & 0.0377 & 0.0762 & 0.6713 & 0.0337 & 0.0329 & 0.0307 & 0.2929 \\
&                     & Alpha360  & 0.0061 & -0.0096 & -0.0032 & -0.0241 & 0.0153 & 0.0231 & 0.0378 & 0.2847 \\
&                     & AlphaAgent & 0.0298 & 0.0244 & 0.0383 & 0.3319 & 0.0386 & 0.0336 & 0.0437 & 0.4411 \\
&                     & RiskMiner & 0.0412 & 0.0356 & -0.0390 & -0.5020 & 0.0414 & 0.0405 & 0.0429 & 0.5726 \\
&                     & Ours & \textbf{0.0420} & \textbf{0.0395} & \textbf{0.0822} & \textbf{0.9397} & \textbf{0.0422} & \textbf{0.0408} & \textbf{0.0737} & \textbf{0.8103} \\
\cmidrule(lr){2-11}
& \multirow{3}{*}{30} & Alpha158  & 0.0028 & 0.0103 & -0.0508 & -0.3716 &  0.0036 & 0.0036 & -0.0122 & -0.1083 \\
&                     & Alpha360  & -0.0539 & -0.0702 & -0.0992 & -0.5688 & 0.0235 & 0.0340 & -0.0136 & -0.1553 \\
&                     & AlphaAgent & 0.0372 & 0.0339 & 0.0412 & 0.3714 & 0.0363 & 0.0344 & 0.0615 & 0.7084 \\
&                     & RiskMiner & \textbf{0.0501} & \textbf{0.0453} & 0.0104 & 0.1151 & 0.0407 & 0.0395 & -0.0332 & -0.3832 \\
&                     & Ours  & 0.0417 & 0.0401 & \textbf{0.0826} & \textbf{1.0312} & \textbf{0.0423} & \textbf{0.0424} & \textbf{0.1315} & \textbf{1.4307} \\
\midrule
\multirow{6}{*}{CSI1000}
& \multirow{3}{*}{10} & Alpha158  & 0.0610 & 0.0627 & 0.0316 & 0.3919 & 0.0520 & 0.0567 & 0.5877 & 1.5969 \\
&                     & Alpha360  & 0.0636 & 0.0642 & 0.0798 & 0.9242 & 0.0550 & 0.0608 & 0.0689 & 0.7576 \\
&                     & AlphaAgent & 0.0726 & 0.0718 & 0.0759 & 0.7111 & \textbf{0.0752} & \textbf{0.0659} & 0.1087 & 1.2615 \\
&                     & RiskMiner & 0.0752 & 0.0708 & 0.0895 & 0.8997 & 0.0712 & 0.0635 & 0.0709 & 0.7872 \\
&                     & Ours & \textbf{0.0804} & \textbf{0.0729} & \textbf{0.1393} & \textbf{1.3577} & 0.0662 & 0.0618 & \textbf{0.1204} & \textbf{1.2868} \\
\cmidrule(lr){2-11}
& \multirow{3}{*}{30} & Alpha158  & 0.0490 & 0.0654 & 0.0343 & 0.5068 & 0.0380 & 0.0373 & 0.0545 & 0.7720 \\
&                     & Alpha360  & 0.0420 & 0.0567 & 0.0171 & 0.2692 & 0.0532 & 0.0456 & 0.0887 & 1.2959 \\
&                     & AlphaAgent & 0.0778 & 0.0692 & 0.0756 & 0.9194 & 0.0713 & 0.0660 & 0.0871 & 1.0454 \\
&                     & RiskMiner & 0.0701 & 0.0686 & 0.1022 & 1.1738 & 0.0722 & 0.0651 & 0.0843 & 1.0918 \\
&                     & Ours & \textbf{0.0793} & \textbf{0.0723} & \textbf{0.1326} & \textbf{1.2598} & \textbf{0.0738} & \textbf{0.0710} & \textbf{0.1696} & \textbf{1.5695} \\
\bottomrule
\end{tabular}
\caption{Predictive performance comparison between our framework and other baselines. For AlphaAgent, RiskMiner and our framework, the size of alpha sets is fixed at 100.}
\label{tab:additional_experimental_results}
\end{table}

\paragraph{Annualized Excess Return (AER)} 
The AER measures the simple arithmetic average rate of excess return per year generated by a portfolio against a market benchmark. For our long-only strategy, at each rebalancing period $t$, we select the top-$k$ assets with the highest predicted returns $f_{i,t}$. The portfolio's performance is based on the realized excess returns of these assets. Let $r^e_{i,t+1}$ denote the realized excess return of asset $i$ over the market benchmark for the period from $t$ to $t+1$. Assuming the $k$ selected assets are equally weighted, the portfolio's excess return for the period $t+1$, denoted $R_{p,t+1}$, is the average of the individual assets' excess returns: 

\begin{equation} 
R_{p,t+1} = \frac{1}{k} \sum_{s \in \text{TopK}_t} r^e_{s,t+1} \end{equation} 

where $\text{TopK}_t$ is the set of $k$ assets with the highest predicted returns based on $\mathbf{f}_t$. The AER, calculated as the arithmetic mean of these per-period portfolio excess returns scaled to an annual figure, over $T_p$ portfolio holding periods is then: 

\begin{equation} 
\text{AER} = \left( \frac{1}{T_p} \sum_{j=1}^{T_p} R_{p,j} \right) \times P \label{eq:aer_simple} 
\end{equation} 

where $R_{p,j}$ is the portfolio excess return over the benchmark in the $j$-th holding period, $T_p$ is the total number of holding periods in the backtest, and $P$ is the number of holding periods in a year (e.g., $P=252$ for daily rebalancing, $P=12$ for monthly rebalancing).

\paragraph{Information Ratio (IR)}
The IR measures the risk-adjusted excess return of the portfolio. It is defined as the Annualized Excess Return divided by its annualized volatility:
\begin{equation}
    \text{IR} = \frac{\text{AER}}{\sigma(R_{p}) \sqrt{P}}
    \label{eq:ir_simple_ar}
\end{equation}
where AER is the Annualized Excess Return calculated as per Eq.~\eqref{eq:aer_simple}, and $\sigma(R_p)$ is the standard deviation of the \textit{per-period} portfolio excess returns $R_{p,j}$ over the $T_p$ holding periods. A higher IR indicates better return per unit of risk.

\section{Additional Results}
\label{sec:appendix_additional_results}

\subsection{Comparisons with Other Baselines}
\label{sec:comarison_with_other_baselines}

\begin{table}[t!]
\centering

\begin{tabular}{@{}c|l|cc|cc|cc@{}}
\toprule
\multirow{2.5}{*}{\textbf{Model}} & \multirow{2.5}{*}{\textbf{Method}} & \multicolumn{2}{c|}{\textbf{Alpha Num = 10}} & \multicolumn{2}{c|}{\textbf{Alpha Num = 50}} & \multicolumn{2}{c}{\textbf{Alpha Num = 100}} \\
\cmidrule(lr){3-4} \cmidrule(lr){5-6} \cmidrule(lr){7-8}
& & IC & RankIC & IC & RankIC & IC & RankIC \\
\midrule
\multirow{8}{*}{LightGBM} 
& GP         & -0.0055 & -0.0055 & 0.0032 & 0.0039 & 0.0060 & 0.0072 \\
& DSO        & 0.0022 & 0.0005 & 0.0038 & 0.0044 & 0.0006 & 0.0015\\
& AlphaGen   & 0.0094 & \textbf{0.0106} & 0.0125 & 0.0121 & 0.0126 & \textbf{0.0134} \\
& AlphaForge & 0.0016 & 0.0007 & 0.0086 & 0.0060 & 0.0110 & 0.0094 \\
& CoT        & 0.0014 & 0.0002 & 0.0106 & 0.0104 & 0.0116 & 0.0112 \\
& ToT        & 0.0032 & 0.0029 & 0.0098 & 0.0111 & 0.0102 & 0.0103 \\
& FAMA       & 0.0086 & 0.0080 & 0.0085 & 0.0095 & 0.0119 & 0.0120 \\
\cmidrule(lr){2-8}
& Ours & \textbf{0.0097} & 0.0102 & \textbf{0.0128} & \textbf{0.0129} & \textbf{0.0132} & 0.0130 \\
\midrule
\multirow{8}{*}{MLP} 
& GP         & -0.0063 & -0.0070 & 0.0048 & 0.0051 & 0.0041 & 0.0047 \\
& DSO        & -0.0017 & -0.0018 & 0.0080 & 0.0079 & 0.0031 & 0.0036 \\
& AlphaGen   & 0.0089 & 0.0094 & 0.0124 & 0.0118 & 0.0117 & 0.0112 \\
& AlphaForge & -0.0001 & 0.0000 & 0.0103 & 0.0104 & 0.0070 & 0.0073 \\
& CoT        & 0.0029 & 0.0035 & 0.0115 & 0.0116 & 0.0113 & 0.0114 \\
& ToT        & 0.0029 & 0.0029 & 0.0092 & 0.0096 & 0.0120 & 0.0113 \\
& FAMA       & 0.0079 & 0.0080 & 0.0109 & 0.0112 & 0.0111 & 0.0108 \\
\cmidrule(lr){2-8}
& Ours & \textbf{0.0092} & \textbf{0.0097} & \textbf{0.0130} & \textbf{0.0127} & \textbf{0.0125} & \textbf{0.0122} \\
\bottomrule
\end{tabular}
\caption{The predictive performance alphas mined by different methods on the S\&P500 stock pool.}
\label{tab:experimental_result_sp500}
\end{table}

To provide a comprehensive assessment of our proposed framework, we conduct additional comparisons against four benchmark alpha sets. The first two are the widely adopted handcrafted libraries \textbf{Alpha158} and \textbf{Alpha360}. Alpha158 consists of 158 factors derived from historical price and volume data for the Chinese A-share market, while Alpha360 is a more extensive set of 360 alphas. We construct these benchmark sets based on the information provided on the Qlib platform~\cite{yang2020qlib}. As a third baseline, we include factors generated by \textbf{AlphaAgent}~\cite{tang2025alphaagent}, a modern LLM-driven approach adapted from the RD-Agent~\cite{yang2025r}. In contrast to purely search-based methods, AlphaAgent leverages human knowledge provided via natural language prompts to guide its discovery process. For our experiments, we direct it to generate 100 factors using the conceptual prompt of discovering ``high-quality alpha factors for the Chinese A-share market.'' As a final baseline, we incorporate \textbf{RiskMiner}~\cite{ren2024riskminer}, a state-of-the-art alpha mining method based on a risk-seeking Monte Carlo Tree Search.

We evaluate the predictive performance of two distinct models: LightGBM and a 3-layer Multi-Layer Perceptron (MLP). Each model is trained using features from five separate sets: (i) Alpha158, (ii) Alpha360, (iii) the AlphaAgent-generated factors, (iv) the RiskMiner-generated factors, and (v) the alphas mined by our proposed method.
The comparative results are presented in Table~\ref{tab:additional_experimental_results}.
As illustrated, our method consistently achieves superior predictive performance across the various experimental settings and model architectures considered, underscoring its effectiveness in identifying more potent alpha signals compared to these established benchmarks.

\subsection{Additional Results on the U.S. Stock Market}
\label{sec:appendix_sp500}

To further assess the generalizability and robustness of our proposed framework, we conduct additional experiments on the U.S. stock market, specifically focusing on the S\&P 500 index constituents. This allows us to validate whether the effectiveness of our alpha mining approach extends to different market environments.

\paragraph{Experimental Setup}
We utilize the daily U.S. stock market data provided by the Qlib platform~\cite{yang2020qlib}, which includes five fundamental features: open, high, low, close prices, and trading volume. The stock pool consists of the constituents of the S\&P 500 index. The prediction target is the 10-day forward return. Following a chronological split, the dataset is divided into a training period from 2007/01/01 to 2015/12/31, and a testing period from 2016/01/01 to 2020/10/10. All other experimental settings, including the alpha generation methods, model configurations (LightGBM and MLP), and evaluation metrics (IC and RankIC), remain consistent with those used for the Chinese market experiments described in the main paper.

\paragraph{Results}
Table~\ref{tab:experimental_result_sp500} presents the predictive performance of alphas generated by our method and the baselines on the S\&P 500 dataset. The results demonstrate that our proposed framework maintains a strong competitive advantage in the U.S. market. Across both LightGBM and MLP models and for all alpha set sizes (10, 50, and 100), our method consistently achieves the highest or near-highest IC and RankIC scores. This suggests that the alphas mined by our framework are not only effective in the Chinese market but also possess robust predictive power in the more mature and competitive U.S. market, confirming the broad applicability of our approach.

\subsection{Investigation of Potential Data Leakage in LLMs}
\label{sec:data_leakage_in_llm}

\begin{table}[t!]
\centering

% \resizebox{0.5\textwidth}{!}{
\begin{tabular}{lcccc}
\toprule
LLM & IC & RankIC & IR & RankIR \\
\midrule
Random & 0.0126 & 0.0179 & 0.095 & 0.148 \\
GPT4.1 & 0.0130 & 0.0242 & 0.097 & 0.168 \\
Gemini2.0-flash-lite & 0.0125 & 0.0231 & 0.094 & 0.180 \\
Deepseek-v3-0324 & 0.0122 & 0.0223 & 0.087 & 0.145 \\
\midrule
Ours & \textbf{0.0527} & \textbf{0.0714} & \textbf{0.368} & \textbf{0.467} \\
\bottomrule
\end{tabular}%}
\caption{Investigation of pre-existing knowledge (data leakage) in LLMs. We compare the performance of alphas generated by directly prompting various LLMs for high-performing formulas against a random baseline and our MCTS-based method. Metrics (IC, RankIC, IR, RankIR) are averages over 10 alphas per method. LLMs (GPT4.1, Gemini2.0-flash-lite, Deepseek-v3) are prompted for high-performing alphas (CSI300, 10-day return). 'Ours' is our method with an alpha set size of 10.}
\label{tab:model-comparison}
\end{table}

A critical concern when employing LLMs for tasks like alpha mining is the potential for data leakage. This refers to the possibility that LLMs' training data might have inadvertently included information about historically well-performing alpha formulas, which could lead to an overestimation of their generative capabilities if they simply recall these formulas.

To investigate this, we design an experiment to explicitly probe whether pre-trained LLMs possess such inherent knowledge. We prompt three distinct LLM backbones—GPT-4.1, Gemini2.0-flash-lite, and Deepseek-v3—to generate alpha formulas specifically anticipated to yield high performance for 10-day forward return prediction on the A-share CSI300 stock pool. For each LLM and the Random baseline, we generate 10 distinct alpha formulas, and the performance metrics reported are averages across these 10 alphas. This setup directly attempts to leverage any pre-existing, ``leaked'' knowledge within the LLMs. The performance of alphas generated under this explicit instruction is compared against two baselines: (1) alphas generated randomly by an LLM (``Random''), and (2) alphas mined by our proposed framework (``Ours'').

As shown in Table~\ref{tab:model-comparison}, the average performance of alphas from LLMs explicitly prompted to generate high-performing expressions is not significantly different from the Random baseline. All these direct LLM generation approaches yield substantially lower performance across all metrics compared to our framework. These results suggest that the tested LLMs do not inherently leverage leaked knowledge of superior alpha formulas for this task, thereby alleviating data leakage concerns and highlighting the efficacy of our alpha mining framework.

\subsection{Sensitivity Analysis of LLM Backbone Choice}
\label{sec:llm_sensitivity}

We evaluate the impact of different LLM backbones on our framework. We test several models from different series: GPT-4.1, Gemini-2.0-flash-lite, Gemini-2.5-flash-lite-preview-06-17, Deepseek-v3, and Qwen3-235b-a22b-2507. Experiments are conducted on the CSI300 stock pool, predicting 10-day forward returns with an alpha set size of 50. Results are detailed in Table \ref{tab:llm_sensitivity}.
The findings reveal that the choice of LLM backbone leads to notable variations in the performance characteristics of the generated alpha sets. While each LLM may emphasize different aspects of alpha quality, our framework, when leveraging these advanced models, generally demonstrates performance that is competitive with or surpasses the best results of baselines.

\begin{table}[!htbp]
    \centering
    
    \begin{tabular}{@{}l cccc cccc@{}}
        \toprule
        \multirow{2.5}{*}{LLM} & \multicolumn{4}{c}{LightGBM} & \multicolumn{4}{c}{MLP} \\
        \cmidrule(lr){2-5} \cmidrule(lr){6-9}
        & IC & RankIC & AR & IR & IC & RankIC & AR & IR \\
        \midrule
        Best of Baselines & 
        0.0388 & 0.0378 & 0.0680 & 0.7470 & 
        0.0380 & 0.0374 & 0.0805 & 0.8414 \\
        \midrule
        GPT4.1 &  
        \textbf{0.0399} & \textbf{0.0394} & 0.0717 & 0.8283 & 
        \textbf{0.0436} & \textbf{0.0425} & 0.0661 & 0.7075 \\
        
        Gemini-2.0-flash-lite & 
        0.0376 & 0.0363 & 0.1037 & 1.1096 & 
        0.0389 & 0.0372 & \textbf{0.1734} & \textbf{1.9801} \\
        
        Gemini-2.5-flash-lite-preview-06-17 & 0.0396 & 0.0390 & \textbf{0.1075} & 1.1885 & 0.0375 & 0.0373 & 0.1225 & 1.3898 \\
        
        Deepseek-v3 & 
        0.0388 & 0.0386 & 0.1064 & \textbf{1.2144} & 
        0.0409 & 0.0401 & 0.1202 & 1.3732 \\
        
        Qwen3-235b-a22b-2507 & 0.0382 & 0.0372 & 0.0614 & 0.7769 & 0.0386 & 0.0380 & 0.1101 & 1.3177 \\
        \bottomrule
    \end{tabular}
    \caption{Performance comparison of our framework using different LLM backbones on the CSI300 dataset (10-day forward returns, alpha set size of 50). Best results are highlighted in \textbf{bold}.}
    \label{tab:llm_sensitivity}
\end{table}

\subsection{Characteristics of Alphas at Varying MCTS Search Depths}
\label{sec:alpha_at_varying_mcts_depth}

In this section, we analysis the characteristics of alphas derived from nodes at different depths within the MCTS search tree. Alphas located at deeper levels of the tree undergo a greater number of refinement iterations. Figure~\ref{fig:mcts_law} illustrates how several key metrics evolve with MCTS depth: specifically, the average formula depth (depth of the formula's tree representation), formula length (number of operators), in-sample (IS) and out-of-sample (OOS) RankIC, and the Overfitting Risk Score.

We observe that as MCTS search depth increases, both the average formula depth and length initially rise before tending to stabilize. This stabilization is largely attributed to the Overfitting Risk Score, which penalizes overly complex formulas more prone to overfitting. Concurrently, both average IS and OOS RankIC exhibit a general upward trend with increasing depth, reflecting the efficacy of the refinement process. Conversely, the generalization gap (the difference between IS and OOS RankIC) tends to widen, indicating a decline in generalization performance as the formulas become more complex. This trend aligns with the behavior of the Overfitting Risk Score, suggesting that while refinement enhances performance, the associated risk of overfitting necessitates careful management, a role our ORS aims to fulfill.

\begin{figure}[t!]
    \centering
    \begin{minipage}[t]{0.48\linewidth}
        \centering
        \includegraphics[width=\linewidth]{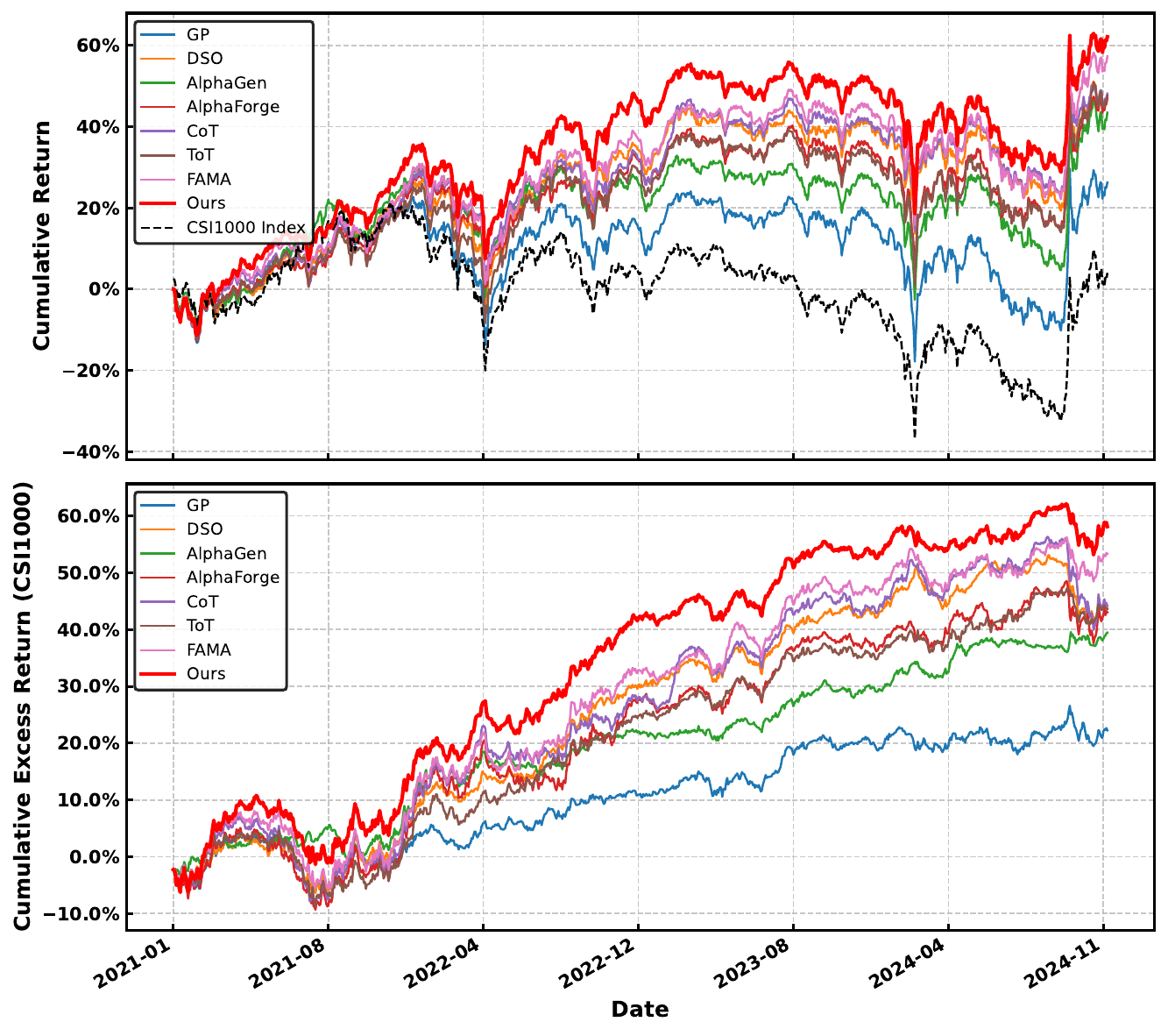}
        \caption*{ (a) LightGBM Model} 
    \end{minipage}
    \hspace{0.02\linewidth}
    \begin{minipage}[t]{0.48\linewidth}
        \centering
        \includegraphics[width=\linewidth]{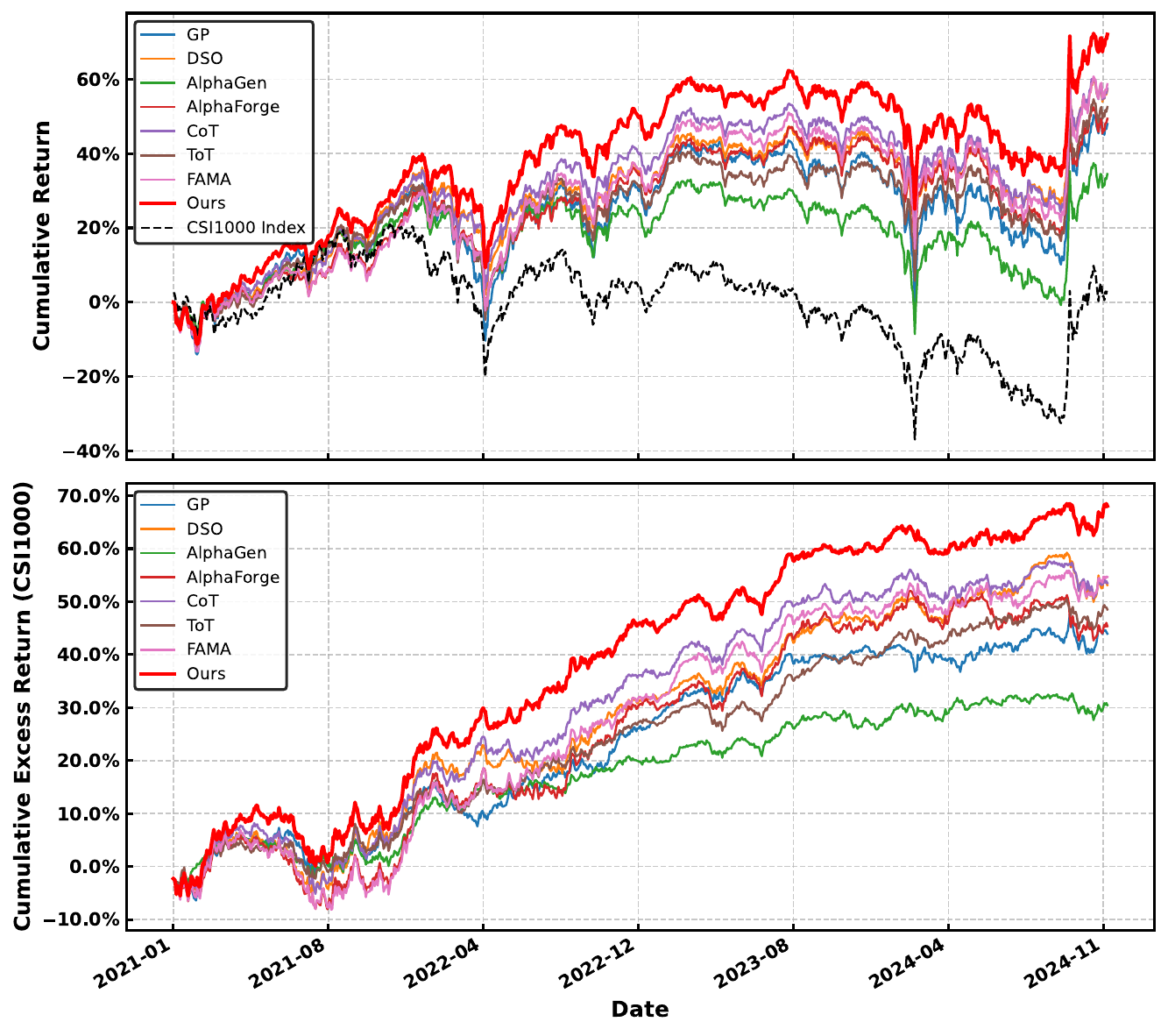}
        \caption*{ (b) MLP Model} 
    \end{minipage}
    \caption{Cumulative return curves of backtest using alphas generated by different methods.}
    \label{fig:backtest_profit}
\end{figure}

\subsection{Cumulative Return Curve Visualizations}
\label{sec:cumu_return_curve}

Figure~\ref{fig:backtest_profit} presents the cumulative return curves derived from backtesting using alphas mined by different methods.
As illustrated, our proposed method consistently demonstrates superior performance, achieving the highest cumulative returns among the evaluated methods.

\subsection{Interpretability of Mined Alpha Formulas}
\label{sec:interpretability_examples}

In quantitative finance, interpretability refers to the ability to connect a mathematical factor to a coherent economic rationale explaining why it might predict cross-sectional returns. An interpretable alpha factor should be grounded in established financial theories or well-documented market phenomena, such as momentum, reversal, liquidity, or volatility effects (e.g., \cite{efficiency1993returns,de1985does,amihud1986asset,french1987expected}). Such grounding is valuable as it provides a plausible economic hypothesis for a factor's potential efficacy. This enhances confidence that the factor captures a genuine market dynamic rather than a spurious, overfitted relationship unlikely to generalize out-of-sample.

\begin{table*}[htbp]
    \centering
    
    \begin{tabularx}{\textwidth}{c >{\raggedright\arraybackslash}X}
        \toprule
        \textbf{Method} & \textbf{Alpha Formula} \\
        \midrule
        \multirow{5}{*}[0pt]{\makecell{Ours}} &
            \begin{minipage}[t]{\linewidth}
            \begin{enumerate}[label=\arabic*., leftmargin=*, itemsep=0.25ex, topsep=0.5ex, parsep=0pt, partopsep=0pt]
                \item $\op{Zscore}(\op{Ma}(\op{close}-\op{vwap},20),30)$
                \item $\op{Std}(\op{Pct}(\op{vwap},20),25)\cdot\op{Sum}(\op{volume},40)/\op{volume}$
                \item $\op{Corr}(\op{close},\op{volume},50)\cdot\op{Zscore}(\op{Ma}(\op{close}-\op{vwap},30),40)$
                \item $\op{Diff}(\op{Ma}(\op{volume},20),3)/\op{Ma}(\op{volume},60)$
                \item $\op{Corr}(\op{Pct}(\op{close},10),\op{Pct}(\op{volume},10),10)\cdot\op{Corr}(\op{Pct}(\op{close},30),$\\
                      $\qquad\op{Pct}(\op{volume},30),30)\cdot\op{Skew}(\op{volume},20)$
                \item $\op{Ma}(\op{Corr}(\op{volume},\op{close},20)\cdot\op{Skew}(\op{high}-\op{low},20),10)$
            \end{enumerate}
            \vspace{1ex}
            \end{minipage} \\ 
        \midrule
        \multirow{2}{*}[0pt]{\makecell{GP}} & 
            \begin{minipage}[t]{\linewidth}
            \begin{enumerate}[label=\arabic*., leftmargin=*, itemsep=0.25ex, topsep=0.5ex, parsep=0pt, partopsep=0pt]
                \item $\op{Add}(\op{Mul}(-0.01,\op{volume}),\op{Log}(\op{Log}(\op{close})))$
                \item $\op{Less}(\op{Cov}(\op{open},\op{Add}(\op{high},\op{Div}(\op{volume},-5)),10),\op{Std}(\op{Log}(\op{close}),50))$
            \end{enumerate}
            \vspace{1ex}
            \end{minipage} \\
        \midrule
        \multirow{2}{*}[0pt]{\makecell{DSO}} & 
            \begin{minipage}[t]{\linewidth}
            \begin{enumerate}[label=\arabic*., leftmargin=*, itemsep=0.25ex, topsep=0.5ex, parsep=0pt, partopsep=0pt]
                \item $\op{Greater}(\op{volume},\op{Med}(\op{Sub}(\op{Ma}(\op{open},10),\op{Med}(\op{Std}(\op{Sign}(\op{close}),10),10)),10))$
                \item $\op{Cov}(\op{Med}(\op{Sign}(\op{vwap}),50),5,20)$
            \end{enumerate}
            \vspace{1ex}
            \end{minipage} \\
        \midrule
        \multirow{2}{*}[0pt]{\makecell{AlphaGen}} & 
            \begin{minipage}[t]{\linewidth}
            \begin{enumerate}[label=\arabic*., leftmargin=*, itemsep=0.25ex, topsep=0.5ex, parsep=0pt, partopsep=0pt]
                \item $\op{Corr}(\op{Rank}(\op{Diff}(\op{Greater}(2.0,\op{volume}),50),10),\op{close},20)$
                \item $\op{Ma}(\op{Greater}(\op{Std}(\op{Less}(0.01,\op{Less}(\op{Div}(\op{Log}(\op{high}),-2.0),-30)),1),-30),5)$
            \end{enumerate}
            \vspace{1ex}
            \end{minipage} \\
        \midrule
        \multirow{2}{*}[0pt]{\makecell{AlphaForge}} & 
            \begin{minipage}[t]{\linewidth}
            \begin{enumerate}[label=\arabic*., leftmargin=*, itemsep=0.25ex, topsep=0.5ex, parsep=0pt, partopsep=0pt]
                \item $1/(1/(\op{Diff}(\op{Sin}(1/(\op{Cos}(((0.01+\op{Sin}(\op{Tanh}(\op{high})))/30)))),30)))$
                \item $(|(\op{Cos}|(\op{Tanh}(\op{Sin}(\op{Diff}(\op{Sin}(30+\op{low}),20)))|))+30.0|)^2$
            \end{enumerate}
            \vspace{1ex}
            \end{minipage} \\
        \bottomrule
    \end{tabularx}
\caption{Examples of alpha formulas mined by our framework and other non-LLM-based baselines. Each numbered item represents a distinct alpha formula.}
    \label{tab:alpha_formula_optimized}
\end{table*}

As shown in Table~\ref{tab:alpha_formula_optimized}, the formulas discovered by our framework can be deconstructed into components with plausible economic intuition. For instance, the formula $\op{Zscore}(\op{Ma}(\op{close}-\op{vwap},20),30)$ can be interpreted as a measure of anomalous intraday momentum. It isolates the end-of-day buying or selling pressure ($\op{close}-\op{vwap}$), smooths it into a short-term trend, and then uses a $\op{Zscore}$ to flag statistically significant deviations. Such abnormal pressure may signal informed trading or strong sentiment, potentially predicting continued returns. Another example, $\op{Std}(\op{Pct}(\op{vwap},20),25)\cdot\op{Sum}(\op{volume},40)/\op{volume}$, combines a measure of recent price volatility with a volume-based reversal signal. This formula identifies assets with high recent volatility but abnormally low current trading activity—a potential signature of trend exhaustion that often precedes price reversals. Other formulas generated by our framework are similarly composed of interpretable concepts like price-volume correlation and volume acceleration (see Table~\ref{tab:alpha_formula_optimized}, formulas 3-6).

In stark contrast, formulas from the baselines often exhibit fundamental issues that undermine their economic plausibility. A primary concern is dimensional inconsistency. For example, the GP formula $\op{Add}(\op{Mul}(-0.01,\op{volume}),\op{Log}(\op{Log}(\op{close})))$ attempts to add a term based on trading volume (unit: shares) to a transformed price term (unit: log-currency). This is analogous to adding mass to length, an operation that lacks a coherent real-world interpretation. Other methods produce expressions like $\op{Std}(\op{Less}(...))$, which calculates the standard deviation of a binary true/false series, a statistically questionable operation. The challenge is most acute for AlphaForge, whose unconstrained search yields formulas like $1/(1/(\op{Diff}(\op{Sin}(1/(\op{Cos}(...))))))$. The nested trigonometric functions applied to financial data are mathematically valid but have no clear basis in economic theory.

While such formulas might fit the training data, their lack of a plausible economic basis makes them resemble ``black boxes." This is a critical drawback, as formulas that violate basic principles like dimensional consistency are at high risk of being spurious artifacts of overfitting. Consequently, they are less likely to be trusted by practitioners and integrated into investment strategies.

\subsection{Sensitivity Analysis of Key Framework Hyperparameters}
\label{sec:sensitivity_analysis_of_framework_param}

\begin{figure}[ht]
  \centering
  \begin{subfigure}[b]{0.48\textwidth}
    \centering
    \includegraphics[width=\textwidth]{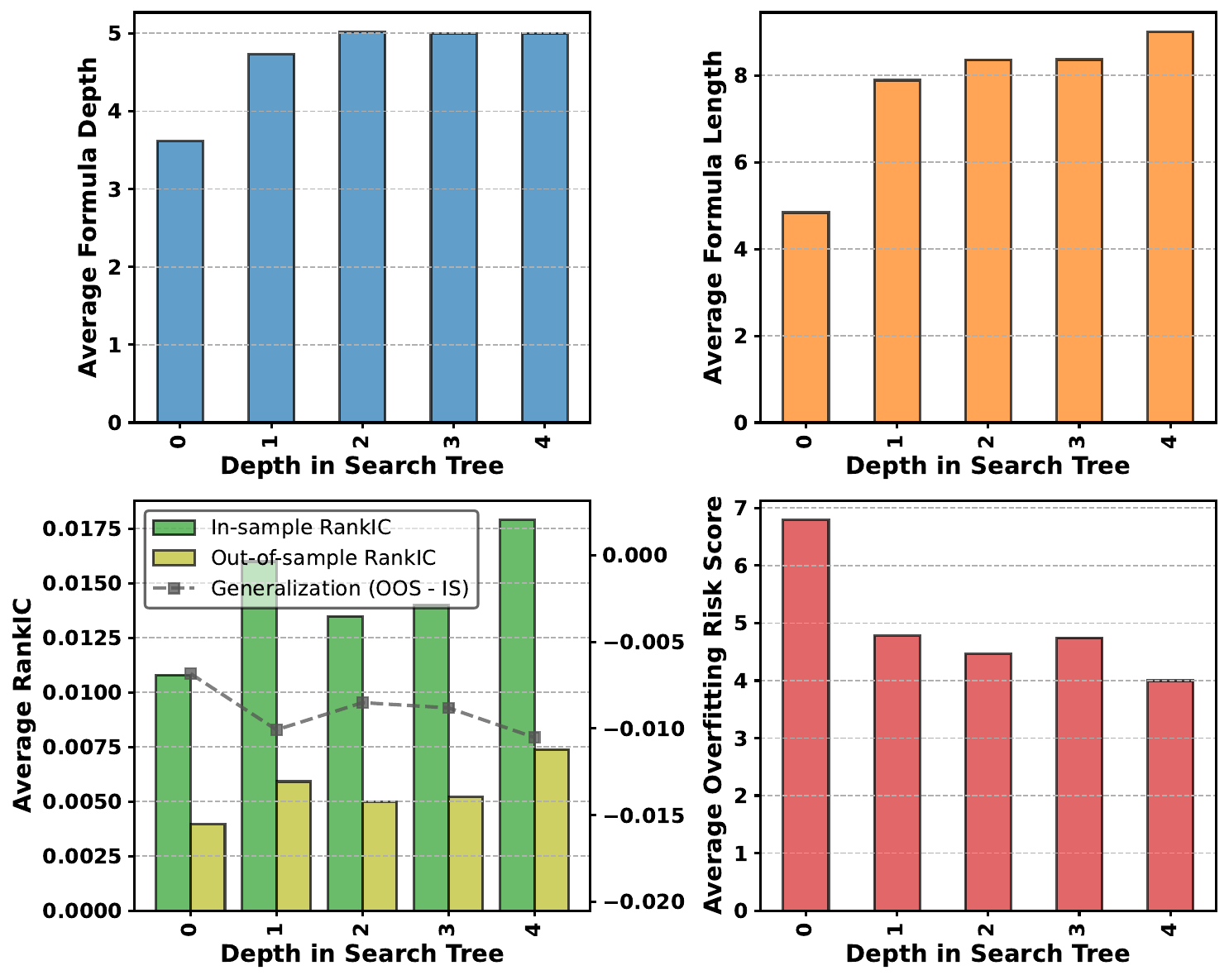}
    \caption{Evolution of Alpha Characteristics with MCTS Search Depth.}
    \label{fig:mcts_law}
  \end{subfigure}
  \hfill
  \begin{subfigure}[b]{0.48\textwidth}
    \centering
    \includegraphics[width=\textwidth]{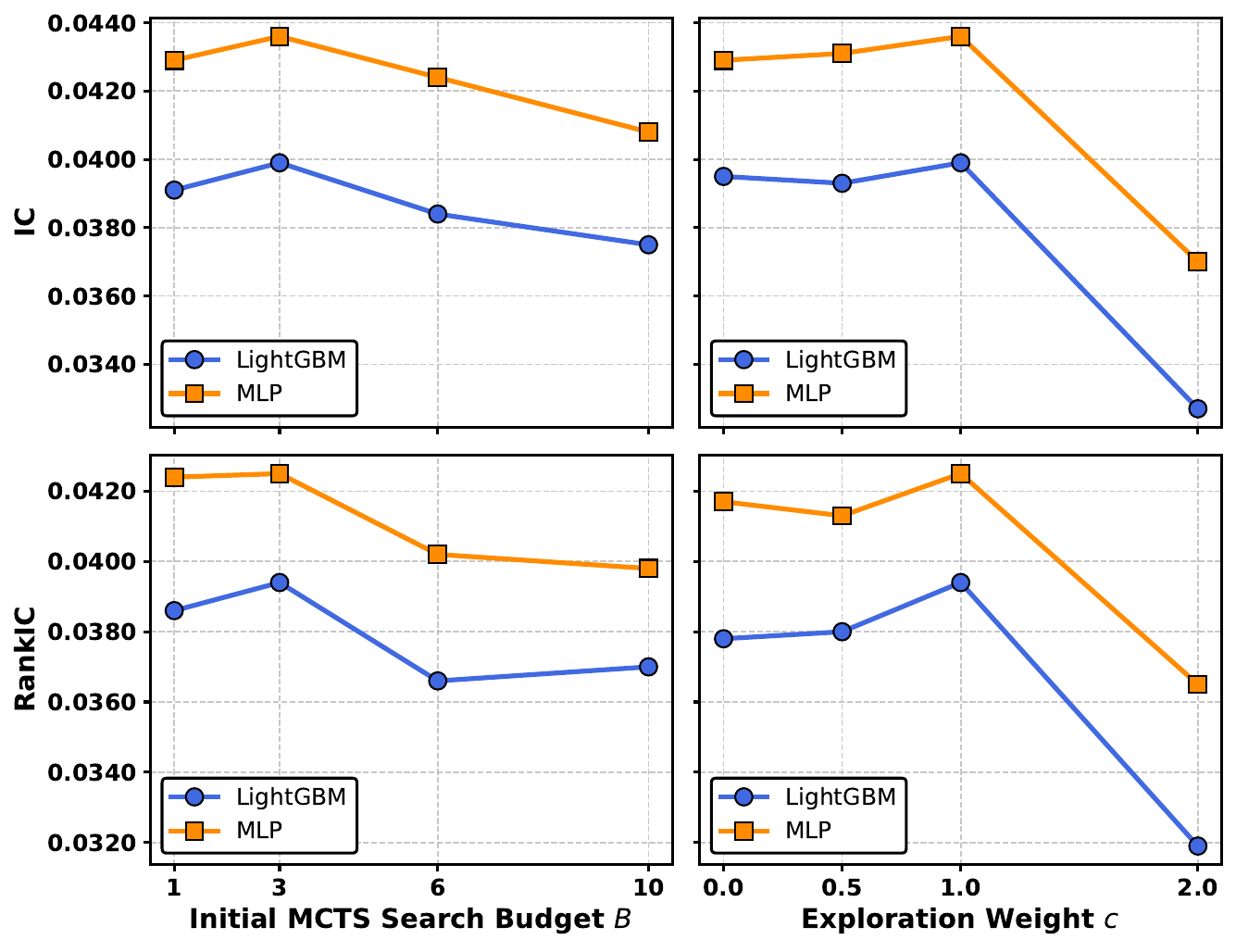}
    \caption{Impact of varying the initial MCTS search budget $B$ and the UCT exploration weight $c$ on the predictive performance.}
    \label{fig:param_sensi_analysis}
  \end{subfigure}

  \caption{Analysis of MCTS search characteristics and parameter sensitivity.}
  \label{fig:mcts_combined}
\end{figure}

In this section, we investigate the sensitivity of our framework to two key hyperparameters: the initial MCTS search budget $B$ (i.e., the initial target search number of MCTS) and the UCT exploration weight $c$.
We conduct experiments using constituents of the CSI300 Index as the stock pool, targeting 10-day forward returns. The alpha set size is set to 50. Figure~\ref{fig:param_sensi_analysis} illustrates the model's performance under various settings for these parameters.

For the initial MCTS search budget $B$, a value of 3 yields the best performance. A larger $B$ can lead to excessive exploration of unpromising MCTS trees, consuming more computational resources without proportional gains. Conversely, a budget that is too small may result in insufficient exploration of potentially promising trees, thereby diminishing search efficacy.

Regarding the UCT exploration weight $c$, our findings suggest that an excessively large value is detrimental. A higher $c$ biases the search towards more uniform exploration (randomness), potentially squandering refinement opportunities on underperforming alpha candidates.

\begin{table}[htbp]
  \centering
  
  \begin{tabular}{l r c r r r r r}
    \toprule
    \textbf{Method} & \textbf{Time (h)} & \textbf{GPU} & \textbf{Input Tokens} & \textbf{Output Tokens} & \multicolumn{3}{c}{\textbf{Cost (\$)}} \\
    \cmidrule(l){6-8}
     & & & & & \textbf{Server} & \textbf{API} & \textbf{Total} \\
    \midrule
    GP          & 28.30 & \checkmark & 0 & 0 & 87.051 & 0.000 & 87.051 \\
    DSO         & 8.70  & \checkmark & 0 & 0 & 26.761 & 0.000 & 26.761 \\
    AlphaGen    & 13.80 & \checkmark & 0 & 0 & 42.449 & 0.000 & 42.449 \\
    AlphaForge  & 0.42  & \checkmark & 0 & 0 & 1.292  & 0.000 & 1.292 \\
    \midrule
    CoT         & 1.42  & \crossmark & 6,110,411 & 1,143,407 & 2.939 & 21.368 & 24.307 \\
    ToT         & 1.73  & \crossmark & 7,072,398 & 1,800,932 & 3.581 & 28.552 & 32.133 \\
    FAMA        & 1.81  & \crossmark & 6,866,434 & 1,198,739 & 3.747 & 23.323 & 27.069 \\
    Ours (GPT4.1)  & 2.76  & \crossmark & 21,391,823 & 3,238,154 & 5.713 & 68.689 & 74.402 \\
    Ours (Gemini-2.0-flash-lite) & 1.94  & \crossmark & 24,175,761 & 5,692,047 & 4.016 & 3.521 & 7.537 \\
    Ours (Deepseek-v3) & 6.91 & \crossmark & 27,291,214 & 5,257,428 & 14.304 & 7.931 & 22.235 \\
    \bottomrule
  \end{tabular}
  \caption{Comparison of single-experiment costs and resource usage across different methods. Time is measured in hours. GPU usage is indicated by a checkmark (\checkmark) for required or a cross (\crossmark) for not required. Costs are presented in USD.}
  \label{tab:cost_comparison}
\end{table}

\subsection{Cost Estimation Details}
\label{sec:cost_estimation}

To provide a standardized and equitable basis for comparing the efficiency of different alpha discovery methods, we conduct a cost-performance analysis. This appendix details the methodology we use to estimate the monetary cost of a single experimental run for each method, as presented in Table~\ref{tab:cost_comparison}. The cost estimation is based on a representative experimental setting: generating an alpha set of 100 alphas on the CSI 300 index.

The total cost for a single run comprises two primary components: server computational cost and LLM API cost.

\paragraph{Server Computational Cost}
The server cost is a function of the experiment's runtime and the required hardware. We reference public cloud computing prices from Amazon Web Services (AWS) for our calculations.
\begin{itemize}
    \item \textbf{CPU-only Server:} For methods that do not require a GPU (all LLM-based methods), we use a 48-core CPU server, priced at \$2.07 per hour.
    \item \textbf{GPU-enabled Server:} For methods requiring GPU acceleration (all non-LLM-based methods), we use a server equipped with an RTX 3080Ti GPU. The total server cost combines the base CPU server price with the GPU price of \$1.006 per hour, for a total hourly rate of \$3.076.
\end{itemize}
The server cost is calculated as: $\text{Cost}_{\text{Server}} = \text{Runtime (h)} \times \text{Hourly Rate (\$/h)}$.

\paragraph{LLM API Cost}
The API cost applies only to LLM-based methods and depends on the number of input and output tokens processed. We use the official pricing for each model. The costs, specified in USD per million tokens for input and output respectively, are as follows:
\begin{itemize}
    \item \textbf{GPT-4.1}: (\$2.00, \$8.00)
    \item \textbf{Gemini-2.0-flash-lite}: (\$0.075, \$0.30)
    \item \textbf{Deepseek-v3}: (\$0.071, \$1.14)
\end{itemize}
The API cost is calculated as: $\text{Cost}_{\text{API}} = (\frac{\text{Input Tokens}}{10^6} \times \text{Rate}_{\text{Input}}) + (\frac{\text{Output Tokens}}{10^6} \times \text{Rate}_{\text{Output}})$.

\paragraph{Estimation Considerations}
It is important to acknowledge two caveats in our estimation:
\begin{itemize}
    \item The cost for non-LLM-based methods is likely an underestimation. Real-world quantitative finance applications often involve higher-frequency data, which would significantly increase the time and computational resources for alpha factor calculation and backtesting.
    \item Conversely, the API cost for LLM-based methods may be a slight overestimation, as we do not account for potential cost reductions from API-level caching, where repeated input tokens can be priced lower.
\end{itemize}

The final total cost in Table~\ref{tab:cost_comparison} is the sum of the server and API costs, offering a unified metric to evaluate search efficiency in terms of performance per dollar.

\begin{table}[!htbp]
\centering

\resizebox{\textwidth}{!}{
\begin{tabular}{@{}c|l|cccc|cccc|cccc@{}}
\toprule
\multirow{2.5}{*}{\textbf{Model}} & \multirow{2.5}{*}{\textbf{Method}} & \multicolumn{4}{c|}{\textbf{Alpha Num = 10}} & \multicolumn{4}{c|}{\textbf{Alpha Num = 50}} & \multicolumn{4}{c}{\textbf{Alpha Num = 100}} \\
\cmidrule(lr){3-6} \cmidrule(lr){7-10} \cmidrule(lr){11-14}
& & IC & RankIC & AR & IR & IC & RankIC & AR & IR & IC & RankIC & AR & IR \\
\midrule
\multirow{4}{*}{LightGBM} 
& CoT        & 0.0201 & 0.0206 & 0.0549 & 0.6713 & 0.0314 & 0.0289 & 0.0531 & 0.5939 & 0.0373 & 0.0352 & 0.0321 & 0.3992 \\
& ToT        & 0.0269 & 0.0267 & 0.0387 & 0.4335 & 0.0261 & 0.0240 & 0.0695 & 0.7544 & 0.0358 & 0.0332 & 0.0620 & 0.6872 \\
& FAMA       & 0.0210 & 0.0206 & 0.0175 & 0.2103 & 0.0243 & 0.0227 & 0.0593 & 0.7470 & 0.0366 & 0.0345 & 0.0731 & 0.8009 \\
\cmidrule(lr){2-14}
& Ours & \textbf{0.0386} & \textbf{0.0364} & \textbf{0.0668} & \textbf{0.7485} & \textbf{0.0399} & \textbf{0.0394} & \textbf{0.0717} & \textbf{0.8283} & \textbf{0.0420} & \textbf{0.0395} & \textbf{0.0822} & \textbf{0.9397} \\
\midrule
\multirow{4}{*}{MLP} 
& CoT        & 0.0207 & 0.0209 & 0.0608 & 0.7030 & 0.0304 & 0.0290 & 0.0566 & 0.6779 & 0.0367 & 0.0334 & 0.0343 & 0.4163 \\
& ToT        & 0.0281 & 0.0278 & 0.0625 & 0.7185 & 0.0299 & 0.0291 & 0.0571 & 0.6340 & 0.0323 & 0.0316 & 0.0340 & 0.3770 \\
& FAMA       & 0.0243 & 0.0227 & 0.0593 & 0.7470 & 0.0292 & 0.0281 & 0.0585 & 0.6868 & 0.0378 & 0.0373 & \textbf{0.0845} & \textbf{0.9401} \\
\cmidrule(lr){2-14}
& Ours & \textbf{0.0411} & \textbf{0.0406} & \textbf{0.0741} & \textbf{0.8186} & \textbf{0.0436} & \textbf{0.0425} & \textbf{0.0661} & \textbf{0.7075} & \textbf{0.0422} & \textbf{0.0408} & 0.0737 & 0.8103 \\
\bottomrule
\end{tabular}}
\caption{Predictive performance of LightGBM and MLP models trained on alphas mined by different methods. The experiment is conducted on the CSI 300 stock pool with a 10-day return prediction horizon.}
\label{tab:equal_cost_comparison}
\end{table}

\paragraph{Comparison under Equal API Cost}
To further scrutinize the search efficiency of the LLM-based methods, we conduct an additional comparison under an equal-cost constraint. For the baseline methods, we incrementally increase the number of generated alphas, evaluating performance at each 1,000-alpha interval until their total API cost matches that of our method. We report the results from the evaluation interval that achieved the highest IC. As shown in Table~\ref{tab:equal_cost_comparison}, our method still achieves superior performance even when baselines are allocated an equivalent API budget. To match this budget, the baseline methods require generating a larger volume of alphas, and their runtimes exceed that of our method. This analysis suggests that our framework demonstrates high search efficiency across the dimensions of search count, token consumption, and runtime.

\section{Full Experimental Results}
\label{sec:full_exp_result}

In this section, we provide detailed experimental results for our proposed method and all other baselines.
Table~\ref{tab:experimental_result_csi300_lgb} and \ref{tab:experimental_result_csi300_mlp} illustrate the experimental results for the LightGBM and MLP models on the CSI300 stock pool.
Similarly, Table~\ref{tab:experimental_result_csi1000_lgb} and \ref{tab:experimental_result_csi1000_mlp} present the results for these two models on the CSI1000 stock pool.

\begin{table}[!htbp]
\centering

\resizebox{\textwidth}{!}{
\begin{tabular}{@{}c|l|cccc|cccc|cccc@{}}
\toprule
\multirow{2.5}{*}{\textbf{$\Delta T$}} & \multirow{2.5}{*}{\textbf{Method}} & \multicolumn{4}{c|}{\textbf{Alpha Num = 10}} & \multicolumn{4}{c|}{\textbf{Alpha Num = 50}} & \multicolumn{4}{c}{\textbf{Alpha Num = 100}} \\
\cmidrule(lr){3-6} \cmidrule(lr){7-10} \cmidrule(lr){11-14}
& & IC & RankIC & AR & IR & IC & RankIC & AR & IR & IC & RankIC & AR & IR \\
\midrule
\multirow{8}{*}{10} 
& GP         & 0.0165 & 0.0148 & 0.0138 & 0.1717 & 0.0248 & 0.0246 & 0.0570 & 0.6777 & 0.0319 & 0.0288 & 0.0716 & 0.7222 \\
& DSO        & 0.0079 & 0.0090 & 0.0434 & 0.6125 & 0.0174 & 0.0178 & -0.0569 & -0.6460 & 0.0246 & 0.0247 & -0.0090 & -0.1069\\
& AlphaGen   & 0.0443 & 0.0411 & 0.0119 & 0.1335 & 0.0388 & 0.0378 & -0.0342 & -0.3473 & \textbf{0.0446} & \textbf{0.0416} & 0.0094 & 0.1091 \\
& AlphaForge & \textbf{0.0521} & \textbf{0.0518} & 0.0231 & 0.2653 & 0.0293 & 0.0271 & 0.0252 & 0.3005 & 0.0407 & 0.0387 & -0.0253 & -0.3004 \\
& CoT        & 0.0201 & 0.0206 & 0.0549 & 0.6713 & 0.0240 & 0.0224 & 0.0600 & 0.6683 & 0.0237 & 0.0225 & 0.0600 & 0.6681 \\
& ToT        & 0.0269 & 0.0267 & 0.0387 & 0.4335 & 0.0256 & 0.0242 & 0.0680 & 0.7365 & 0.0358 & 0.0332 & 0.0620 & 0.6872 \\
& FAMA       & 0.0210 & 0.0206 & 0.0175 & 0.2103 & 0.0243 & 0.0227 & 0.0593 & 0.7470 & 0.0292 & 0.0281 & 0.0585 & 0.6868 \\
\cmidrule(lr){2-14}
& Ours & 0.0386 & 0.0364 & \textbf{0.0668} & \textbf{0.7485} & \textbf{0.0399} & \textbf{0.0394} & \textbf{0.0717} & \textbf{0.8283} & 0.0420 & 0.0395 & \textbf{0.0822} & \textbf{0.9397} \\
\midrule
\multirow{8}{*}{30} 
& GP         & 0.0145 & 0.0118 & 0.0322 & 0.3857 & 0.0248 & 0.0246 & 0.0570 & 0.6777 & 0.0319 & 0.0288 & 0.0716 & 0.7222 \\
& DSO        & 0.0079 & 0.0090 & -0.0050 & -0.0682 & 0.0174 & 0.0178 & 0.0156 & 0.1735 & 0.0246 & 0.0247 & 0.0185 & 0.2039 \\
& AlphaGen   & 0.0299 & 0.0280 & 0.0597 & 0.7792 & 0.0322 & 0.0286 & 0.0150 & 0.1540 & 0.0401 & 0.0339 & 0.0548 & 0.4662 \\
& AlphaForge & 0.0326 & 0.0322 & 0.0981 & 0.9455 & 0.0286 & 0.0278 & 0.0551 & 0.7341 & 0.0339 & 0.0315 & 0.0164 & 0.2176 \\
& CoT        & 0.0200 & 0.0186 & 0.0774 & 0.9554 & 0.0274 & 0.0250 & 0.0574 & 0.6105 & 0.0278 & 0.0247 & 0.0468 & 0.5037 \\
& ToT        & 0.0232 & 0.0247 & 0.1064 & 1.3221 & 0.0291 & 0.0285 & 0.0632 & 0.7982 & 0.0348 & 0.0312 & 0.0587 & 0.7037 \\
& FAMA       & 0.0298 & 0.0307 & 0.1050 & 1.1609 & 0.0285 & 0.0298 & 0.0803 & 0.9502 & 0.0301 & 0.0297 & \textbf{0.0934} & \textbf{1.1629} \\
\cmidrule(lr){2-14}
& Ours & \textbf{0.0334} & \textbf{0.0334} & \textbf{0.1129} & \textbf{1.3286} & \textbf{0.0352} & \textbf{0.0340} & \textbf{0.0886} & \textbf{1.1299} & \textbf{0.0417} & \textbf{0.0401} & 0.0826 & 1.0312 \\
\bottomrule
\end{tabular}}
\caption{The predictive performance of LightGBM model trained on alphas mined by different methods on the CSI300 stock pool.}
\label{tab:experimental_result_csi300_lgb}
\end{table}

\begin{table}[!htbp]
\centering

\resizebox{\textwidth}{!}{
\begin{tabular}{@{}c|l|cccc|cccc|cccc@{}}
\toprule
\multirow{2.5}{*}{\textbf{$\Delta T$}} & \multirow{2.5}{*}{\textbf{Method}} & \multicolumn{4}{c|}{\textbf{Alpha Num = 10}} & \multicolumn{4}{c|}{\textbf{Alpha Num = 50}} & \multicolumn{4}{c}{\textbf{Alpha Num = 100}} \\
\cmidrule(lr){3-6} \cmidrule(lr){7-10} \cmidrule(lr){11-14}
& & IC & RankIC & AR & IR & IC & RankIC & AR & IR & IC & RankIC & AR & IR \\
\midrule
\multirow{8}{*}{10} 
& GP         & 0.0124 & 0.0120 & -0.0212 & -0.2240 & 0.0257 & 0.0233 & 0.0769 & 0.8324 & 0.0277 & 0.0266 & 0.0345 & 0.3967 \\
& DSO        & 0.0085 & 0.0082 & 0.0485 & 0.5849 & 0.0062 & 0.0052 & -0.0263 & -0.2623 & 0.0256 & 0.0245 & 0.0019 & 0.0230 \\
& AlphaGen   & 0.0419 & 0.0418 & 0.0725 & 0.8199 & 0.0380 & 0.0374 & 0.0524 & 0.6187 & 0.0311 & 0.0313 & 0.0295 & 0.3044 \\
& AlphaForge & \textbf{0.0549} & \textbf{0.0554} & \textbf{0.0802} & \textbf{0.8397} & 0.0232 & 0.0207 & 0.0668 & 0.7509 & 0.0408 & 0.0395 & 0.0220 & 0.2555 \\
& CoT        & 0.0207 & 0.0209 & 0.0608 & 0.7030 & 0.0260 & 0.0260 & 0.0500 & 0.5632 & 0.0263 & 0.0257 & 0.0592 & 0.6627 \\
& ToT        & 0.0281 & 0.0278 & 0.0625 & 0.7185 & 0.0285 & 0.0286 & \textbf{0.0805} & \textbf{0.8414} & 0.0323 & 0.0316 & 0.0340 & 0.3770 \\
& FAMA       & 0.0243 & 0.0227 & 0.0593 & 0.7470 & 0.0292 & 0.0281 & 0.0585 & 0.6868 & 0.0307 & 0.0319 & \textbf{0.1013} & \textbf{1.2209} \\
\cmidrule(lr){2-14}
& Ours & 0.0411 & 0.0406 & 0.0741 & 0.8186 & \textbf{0.0436} & \textbf{0.0425} & 0.0661 & 0.7075 & \textbf{0.0422} & \textbf{0.0408} & 0.0737 & 0.8103 \\
\midrule
\multirow{8}{*}{30} 
& GP         & 0.0149 & 0.0144 & 0.0541 & 0.6154 & 0.0262 & 0.0245 & 0.1422 & 1.5026 & 0.0220 & 0.0194 & 0.0815 & 0.7696 \\
& DSO        & -0.0009 & 0.0017 & -0.0605 & -0.6320 & 0.0067 & 0.0057 & 0.0037 & 0.0378 & 0.0250 & 0.0230 & 0.0919 & 0.9969 \\
& AlphaGen   & 0.0204 & 0.0190 & 0.0417 & 0.4695 & 0.0240 & 0.0257 & 0.0538 & 0.6088 & 0.0415 & 0.0400 & 0.1149 & 1.2440 \\
& AlphaForge & 0.0348 & 0.0355 & 0.1162 & 1.0065 & 0.0283 & 0.0274 & 0.1004 & 1.2230 & 0.0317 & 0.0280 & 0.0732 & 0.9253 \\
& CoT        & 0.0198 & 0.0188 & 0.1050 & 1.2943 & 0.0274 & 0.0264 & \textbf{0.1631} & \textbf{1.6693} & 0.0276 & 0.0250 & 0.1042 & 1.0422 \\
& ToT        & 0.0265 & 0.0262 & 0.0975 & 1.1829 & \textbf{0.0355} & \textbf{0.0353} & 0.1330 & 1.5599 & 0.0337 & 0.0327 & 0.0584 & 0.7347 \\
& FAMA       & 0.0341 & 0.0338 & 0.1173 & 1.3005 & 0.0302 & 0.0300 & 0.0596 & 0.6626 & 0.0331 & 0.0327 & 0.1012 & 1.0366 \\
\cmidrule(lr){2-14}
& Ours & \textbf{0.0361} & \textbf{0.0359} & \textbf{0.1259} & \textbf{1.3210} & 0.0337 & 0.0340 & 0.1235 & 1.3931 & \textbf{0.0423} & \textbf{0.0424} & \textbf{0.1315} & \textbf{1.4307} \\
\bottomrule
\end{tabular}}
\caption{The predictive performance of MLP model trained on alphas mined by different methods on the CSI300 stock pool.}
\label{tab:experimental_result_csi300_mlp}
\end{table}

\begin{table}[!htbp]
\centering

\resizebox{\textwidth}{!}{
\begin{tabular}{@{}c|l|cccc|cccc|cccc@{}}
\toprule
\multirow{2.5}{*}{\textbf{$\Delta T$}} & \multirow{2.5}{*}{\textbf{Method}} & \multicolumn{4}{c|}{\textbf{Alpha Num = 10}} & \multicolumn{4}{c|}{\textbf{Alpha Num = 50}} & \multicolumn{4}{c}{\textbf{Alpha Num = 100}} \\
\cmidrule(lr){3-6} \cmidrule(lr){7-10} \cmidrule(lr){11-14}
& & IC & RankIC & AR & IR & IC & RankIC & AR & IR & IC & RankIC & AR & IR \\
\midrule
\multirow{8}{*}{10} 
& GP         & 0.0621 & 0.0581 & 0.0965 & 0.8339 & 0.0627 & 0.0587 & 0.0956 & 0.8933 & 0.0744 & 0.0646 & 0.1054 & 1.0063 \\
& DSO        & 0.0362 & 0.0368 & 0.0926 & 0.8602 & 0.0492 & 0.0466 & 0.1091 & 1.0195 & 0.0609 & 0.0559 & 0.1137 & 1.0720 \\
& AlphaGen   & 0.0528 & 0.0511 & 0.0770 & 0.7751 & \textbf{0.0828} & \textbf{0.0720} & 0.0403 & 0.4799 & 0.0793 & 0.0693 & 0.0816 & 0.9041 \\
& AlphaForge & \textbf{0.0702} & \textbf{0.0628} & 0.0768 & 0.7029 & 0.0664 & 0.0601 & 0.0881 & 0.7687 & 0.0728 & 0.0631 & 0.0512 & 0.4607 \\
& CoT        & 0.0556 & 0.0498 & 0.0450 & 0.4726 & 0.0638 & 0.0574 & 0.0607 & 0.6113 & 0.0670 & 0.0597 & 0.0602 & 0.5818 \\
& ToT        & 0.0619 & 0.0573 & 0.0993 & 0.9592 & 0.0599 & 0.0558 & 0.0954 & 0.9419 & 0.0654 & 0.0585 & 0.0909 & 0.8889 \\
& FAMA       & 0.0631 & 0.0595 & 0.1053 & 0.9234 & 0.0643 & 0.0596 & 0.1134 & 1.0222 & 0.0647 & 0.0584 & 0.1186 & 1.0708 \\
\cmidrule(lr){2-14}
& Ours & 0.0661 & 0.0603 & \textbf{0.1096} & \textbf{1.0919} & 0.0748 & 0.0677 & \textbf{0.1418} & \textbf{1.3699} & \textbf{0.0804} & \textbf{0.0729} & \textbf{0.1393} & \textbf{1.3577} \\
\midrule
\multirow{8}{*}{30} 
& GP         & \textbf{0.0575} & 0.0522 & 0.0996 & 1.1359 & 0.0674 & 0.0637 & 0.0559 & 0.7088 & 0.0706 & 0.0672 & 0.0496 & 0.6230 \\
& DSO        & 0.0362 & 0.0368 & 0.1181 & 1.1052 & 0.0420 & 0.0397 & 0.1096 & 1.2138 & 0.0609 & 0.0559 & 0.1226 & 1.1488 \\
& AlphaGen   & 0.0511 & 0.0502 & 0.1087 & 1.2971 & 0.0727 & 0.0634 & 0.0991 & \textbf{1.4196} & 0.0685 & 0.0686 & 0.0520 & 0.7458 \\
& AlphaForge & 0.0558 & \textbf{0.0532} & 0.1095 & 0.9708 & 0.0705 & 0.0638 & 0.1081 & 0.9519 & 0.0705 & 0.0611 & 0.0796 & 0.8501 \\
& CoT        & 0.0471 & 0.0438 & 0.1030 & 1.1473 & 0.0693 & 0.0629 & 0.1110 & 0.9757 & 0.0743 & 0.0661 & 0.1118 & 1.0872 \\
& ToT        & 0.0475 & 0.0458 & 0.1340 & 1.4212 & 0.0643 & 0.0576 & 0.1097 & 1.1991 & 0.0758 & 0.0694 & 0.1147 & 1.1127 \\
& FAMA       & 0.0543 & 0.0522 & 0.1136 & 0.9774 & 0.0550 & 0.0523 & 0.1340 & 1.2054 & 0.0557 & 0.0523 & 0.1226 & 1.0912 \\
\cmidrule(lr){2-14}
& Ours & 0.0529 & 0.0506 & \textbf{0.1543} & \textbf{1.6588} & \textbf{0.0741} & \textbf{0.0673} & \textbf{0.1461} & 1.4065 & \textbf{0.0793} & \textbf{0.0723} & \textbf{0.1326} & \textbf{1.2598} \\
\bottomrule
\end{tabular}}
\caption{The predictive performance of LightGBM model trained on alphas mined by different methods on the CSI1000 stock pool.}
\label{tab:experimental_result_csi1000_lgb}
\end{table}

\begin{table}[!htbp]
\centering

\resizebox{\textwidth}{!}{
\begin{tabular}{@{}c|l|cccc|cccc|cccc@{}}
\toprule
\multirow{2.5}{*}{\textbf{$\Delta T$}} & \multirow{2.5}{*}{\textbf{Method}} & \multicolumn{4}{c|}{\textbf{Alpha Num = 10}} & \multicolumn{4}{c|}{\textbf{Alpha Num = 50}} & \multicolumn{4}{c}{\textbf{Alpha Num = 100}} \\
\cmidrule(lr){3-6} \cmidrule(lr){7-10} \cmidrule(lr){11-14}
& & IC & RankIC & AR & IR & IC & RankIC & AR & IR & IC & RankIC & AR & IR \\
\midrule
\multirow{8}{*}{10} 
& GP         & 0.0578 & 0.0555 & 0.0922 & 0.8226 & 0.0606 & 0.0572 & 0.0807 & 0.6590 & 0.0659 & 0.0613 & 0.1031 & 0.9692 \\
& DSO        & 0.0355 & 0.0349 & 0.1043 & 0.9314 & 0.0416 & 0.0405 & 0.0820 & 0.7957 & 0.0502 & 0.0469 & 0.0926 & 0.8907 \\
& AlphaGen   & 0.0462 & 0.0463 & 0.0947 & 0.9076 & 0.0665 & 0.0619 & 0.0445 & 0.5072 & 0.0659 & 0.0597 & 0.0261 & 0.4004 \\
& AlphaForge & \textbf{0.0694} & \textbf{0.0634} & 0.1060 & 0.9916 & 0.0633 & 0.0580 & 0.1086 & 0.9420 & 0.0632 & 0.0606 & 0.0631 & 0.5837 \\
& CoT        & 0.0518 & 0.0480 & 0.0513 & 0.5369 & 0.0585 & 0.0544 & 0.0721 & 0.7412 & 0.0648 & 0.0593 & 0.0706 & 0.6824 \\
& ToT        & 0.0606 & 0.0578 & 0.1118 & 1.0644 & 0.0575 & 0.0558 & 0.0934 & 0.9376 & 0.0637 & \textbf{0.0622} & 0.0971 & 0.9258 \\
& FAMA       & 0.0620 & 0.0591 & 0.1045 & 0.9361 & 0.0611 & 0.0582 & 0.1011 & 0.9539 & 0.0635 & 0.0572 & 0.1041 & 1.0323 \\
\cmidrule(lr){2-14}
& Ours & 0.0609 & 0.0581 & \textbf{0.1286} & \textbf{1.3039} & \textbf{0.0681} & \textbf{0.0622} & \textbf{0.1168} & \textbf{1.1843} & \textbf{0.0662} & 0.0618 & \textbf{0.1204} & \textbf{1.2868} \\
\midrule
\multirow{8}{*}{30} 
& GP         & \textbf{0.0555} & \textbf{0.0517} & 0.1590 & 1.4989 & 0.0653 & 0.0606 & 0.1105 & 1.2125 & 0.0655 & 0.0638 & 0.0977 & 1.0269 \\
& DSO        & 0.0312 & 0.0318 & 0.1171 & 1.0981 & 0.0338 & 0.0326 & 0.1336 & 1.3199 & 0.0542 & 0.0522 & 0.1056 & 0.9626 \\
& AlphaGen   & 0.0505 & 0.0505 & 0.1436 & 1.3123 & 0.0664 & 0.0611 & 0.0766 & 1.0838 & 0.0640 & 0.0622 & 0.0569 & 0.8894 \\
& AlphaForge & 0.0541 & \textbf{0.0517} & 0.1071 & 0.9857 & 0.0642 & 0.0606 & 0.1134 & 1.0138 & 0.0717 & 0.0650 & 0.0667 & 0.6850 \\
& CoT        & 0.0465 & 0.0440 & 0.1176 & 1.2683 & 0.0662 & 0.0620 & 0.1345 & 1.3550 & 0.0698 & 0.0615 & 0.1174 & 1.4363 \\
& ToT        & 0.0466 & 0.0447 & 0.1322 & 1.3771 & 0.0608 & 0.0556 & 0.1220 & 1.4510 & 0.0690 & 0.0633 & 0.1116 & 1.2468 \\
& FAMA       & 0.0532 & 0.0511 & 0.1224 & 1.1036 & 0.0538 & 0.0519 & 0.1374 & 1.2472 & 0.0571 & 0.0530 & 0.1327 & 1.2086 \\
\cmidrule(lr){2-14}
& Ours & 0.0514 & 0.0491 & \textbf{0.1799} & \textbf{1.8021} & \textbf{0.0673} & \textbf{0.0647} & \textbf{0.1710} & \textbf{1.6268} & \textbf{0.0738} & \textbf{0.0710} & \textbf{0.1696} & \textbf{1.5695} \\
\bottomrule
\end{tabular}}
\caption{The predictive performance of MLP model trained on alphas mined by different methods on the CSI1000 stock pool.}
\label{tab:experimental_result_csi1000_mlp}
\end{table}

\section{Limitations}
\label{sec:limitation}

Despite the promising advancements presented by our framework for formulaic alpha mining, several limitations warrant discussion.

First, while our method generates effective alpha formulas, there remains a gap in achieving the same level of novelty and complexity typically found in formulas developed by human experts. The framework may sometimes struggle to produce highly intricate or unconventional alphas.
Second, the diversity of the generated alphas is inherently constrained by the internal knowledge base of LLMs. This can, in turn, limit the breadth of the search space explored compared to certain non-LLM-based methodologies.
Consequently, scaling our approach to extremely large-scale alpha mining tasks, which necessitate the exploration of a vast and diverse alpha landscape, may present challenges.

Addressing these limitations, such as by incorporating mechanisms for enhanced novelty or expanding the effective search space, constitutes important directions for future research.

\section{LLM Agent Prompts}
\label{sec:appendix_prompts}

In our proposed framework, LLMs function as autonomous agents, instrumental in the multifaceted process of alpha formula mining. This includes the generation, iterative refinement, and overfitting risk evaluation of alpha formulas. This section details the core prompts designed to guide the LLM in these pivotal tasks. Each prompt is carefully designed to elicit specific behaviors and outputs from the LLM, ensuring a structured and effective approach to mining alpha formulas.

\subsection{Alpha Portrait Generation Prompt}

As shown in Figure~\ref{fig:portrait_generation_prompt}, the Alpha Portrait Generation Prompt is used to generate an alpha portrait based on information such as available data fields and operators. The alpha portrait is subsequently used to generate the corresponding alpha formula.

\subsection{Alpha Formula Generation Prompt}

As shown in Figure~\ref{fig:formula_generation_prompt}, the Alpha Formula Generation Prompt is used to generate the corresponding alpha formula based on the provided alpha portrait and other relevant information. A formatted alpha formula facilitates correctness verification and the computation of alpha values.

\subsection{Alpha Overfitting Risk Assessment Prompt}
\label{sec:alpha_overfitting_risk_prompt}

As shown in Figure~\ref{fig:overfitting_assessment_prompt}, the Alpha Overfitting Risk Assessment Prompt is used to assess the overfitting risk of an alpha based on its formula and refinement history information. We provide evaluation criteria within the prompt to assist the LLM in conducting a critical assessment.

\subsection{Alpha Refinement Prompt}

As shown in Figure~\ref{fig:refinement_prompt}, the Alpha Refinement Prompt is used to refine the original alpha formula based on refinement suggestions, thereby generating an improved formula.

\begin{figure}[htbp]
    \centering
    \small
    \begin{minipage}{0.95\textwidth}
        \hrulefill
        \vspace{1ex}

        \noindent\textbf{Task Description:}

        \noindent You are a quantitative finance expert specializing in factor-based investing. Please design an alpha factor used in investment strategies according to the following requirements, and then provide the content of the alpha in the required format.

        \vspace{1ex}
        \noindent\textbf{Available Data Fields:}

        \noindent The following data fields are available for use:
\begin{verbatim}
{available_fields}
\end{verbatim}

        \vspace{1ex}
        \noindent\textbf{Available Operators:}

        \noindent The following operators are available for use:
\begin{verbatim}
{available_operators}
\end{verbatim}

        \vspace{1ex}
        \noindent\textbf{Alpha Requirements:}
        \begin{enumerate}
            \item The alpha value should be dimensionless (unitless).
            \item The alpha should incorporate at least two distinct operations from the "Available Operators" list to ensure it has sufficient complexity. Avoid creating overly simplistic alphas.
            \item All look-back windows and other numerical parameters used in the alpha calculation MUST be represented as named parameters in the pseudo-code. These parameter names MUST follow Python naming conventions (e.g., \texttt{lookback\_period}, \texttt{volatility\_window}, \texttt{smoothing\_factor}).
            \item The alpha should have NO MORE than 3 parameters in total.
            \item The pseudo-code should represent the alpha calculation step-by-step, using only the "Available Operators" and clearly defined parameters. Each line in the pseudo-code should represent a single operation.
            \item Use descriptive variable names in the pseudo-code that clearly indicate the data they represent.
            \item When designing alpha expressions, try to avoid including the following sub-expressions:
\begin{verbatim}
{freq_subtrees}
\end{verbatim}
        \end{enumerate}

        \vspace{1ex}
        \noindent\textbf{Formatting Requirements:}

        \noindent The output must be in JSON format with three key-value pairs:
        \begin{enumerate}
            \item \textbf{"name":} A short, descriptive name for the alpha (following Python variable naming style, e.g., \texttt{price\_volatility\_ratio}).
            \item \textbf{"description":} A concise explanation of the alpha's purpose or what it measures. Avoid overly technical language. Focus on the intuition behind the alpha.
            \item \textbf{"pseudo\_code":} A list of strings, where each string is a line of simplified pseudo-code representing a single operation in the alpha calculation. Each line should follow the format: \texttt{variable\_name = op\_name(input=[input1, input2, ...], param=[param1, param2, ...])}, where:
                \begin{itemize}
                    \item \texttt{variable\_name} is the output variable of the operation.
                    \item \texttt{op\_name} is the name of one of the "Available Operators".
                    \item \texttt{input1}, \texttt{input2}, \dots are input variables (either from "Available Data Fields" or previously calculated variables, cannot be of a numeric type).
                    \item \texttt{param1}, \texttt{param2}, \dots are parameter names defined in the alpha requirements.
                \end{itemize}
        \end{enumerate}

        \vspace{1ex}
        \noindent The format example is as follows:

\begin{lstlisting}
{
    "name": "volatility_adjusted_momentum",
    "description": "......",
    "pseudo_code": [......]
}
\end{lstlisting}
        \vspace{1ex}
        \hrulefill
    \end{minipage}
\caption{Alpha Portrait Generation Prompt}
\label{fig:portrait_generation_prompt}
\end{figure}

\begin{figure}[htbp]
    \centering
    \small
    \begin{minipage}{0.95\textwidth}
        \hrulefill
        \vspace{1ex}

        \noindent\textbf{Task Description:}
        \vspace{1ex} % Added for spacing

        Please design a quantitative investment alpha expression according to the following requirements.

        \vspace{1ex}
        \noindent\textbf{Available Data Fields:}
        \begin{itemize}
            \item The following data fields are available for use: \texttt{\{available\_fields\}}
        \end{itemize}

        \vspace{1ex}
        \noindent\textbf{Available Operators:}
         \begin{itemize}
            \item The following operators are available for use: \texttt{\{available\_operators\}}
        \end{itemize}

        \vspace{1ex}
        \noindent\textbf{Alpha Requirements:}
        \begin{itemize}
            \item \texttt{\{alpha\_portrait\_prompt\}}
        \end{itemize}

        \vspace{1ex}
        \noindent\textbf{Formatting Requirements:}
        \begin{enumerate}[label=\arabic*.] % Ensure consistent numbering style
            \item Provide the output in JSON format.
            \item The JSON object should contain two fields: \texttt{"formula"}, and \texttt{"arguments"}.
                \begin{itemize}
                    \item \texttt{"formula"}: Represents the mathematical expression for calculating the alpha.
                    \item \texttt{"arguments"}: Represents the configurable parameters of the alpha.
                \end{itemize}
            \item \texttt{"formula"} is a list of dictionaries. Each dictionary represents a single operation and must contain four keys: \texttt{"name"}, \texttt{"param"}, \texttt{"input"}, and \texttt{"output"}.
                \begin{itemize}
                    \item \texttt{"name"}: The operator's name (a string), which MUST be one of the operators provided in the "Available Operators" section.
                    \item \texttt{"param"}: A list of strings, representing the parameter names for the operator. These parameter names MUST be used as keys in the \texttt{"arguments"} section.
                    \item \texttt{"input"}: A list of strings, representing the input variable names for the operator. These MUST be data fields from the "Available Data Fields" or output variables from previous operations in the \texttt{"formula"}, cannot be of a numeric type.
                    \item \texttt{"output"}: A string, representing the output variable name for the operator. This output can be used as an input for subsequent operations.
                \end{itemize}
            \item \texttt{"arguments"} is a list of dictionaries. Each dictionary represents a set of parameter values for the alpha.
                \begin{itemize}
                    \item The keys of each dictionary in \texttt{"arguments"} MUST correspond exactly to the parameter names defined in the \texttt{"param"} lists of the \texttt{"formula"}.
                    \item The values in each dictionary in \texttt{"arguments"} are the specific numerical values for the parameters.
                \end{itemize}
            \item You may include a maximum of 3 sets of parameters within the \texttt{"arguments"} field.
            \item The parameter value that indicates the length of the lookback window (if applicable) must be within the \texttt{\{window\_range\}} range.
            \item Ensure that the alpha expression is both reasonable and computationally feasible.
            \item Parameter names should be descriptive and follow Python naming conventions (e.g., \texttt{window\_size}, \texttt{lag\_period}, \texttt{smoothing\_factor}). Avoid using single characters or numbers as parameter names.
            \item Refer to the following example:
        \end{enumerate}
        
        \vspace{1ex}
        \begin{lstlisting}
{
    "formula": [......],
    "arguments": [......]
}
        \end{lstlisting}
        \vspace{1ex}
        \hrulefill
    \end{minipage}
\caption{Alpha Formula Generation Prompt}
\label{fig:formula_generation_prompt}
\end{figure}

\begin{figure}[htbp]
    \centering
    \small
    \begin{minipage}{0.95\textwidth}
        \hrulefill
        \vspace{1ex}

        \noindent\textbf{Task: Critical Alpha Overfitting Risk Assessment}

        \noindent Critically evaluate the overfitting risk and generalization potential of the provided quantitative investment alpha, based on its expression and refinement history.
        Your assessment must focus on whether complexity and optimization appear justified or are likely signs of overfitting.

        \vspace{1ex}
        \noindent\textbf{Input:}
        \begin{itemize}
            \item \textbf{Alpha Expression:}
\begin{verbatim}
{alpha_formula}
\end{verbatim}
            \item \textbf{Refinement History:}
\begin{verbatim}
{refinement_history}
\end{verbatim}
        \end{itemize}

        \vspace{1ex}
        \noindent\textbf{Evaluation Criteria:}
        \begin{enumerate}
            \item \textbf{Justified Rationale vs. Complexity:}
                \begin{itemize}
                    \item[] \textbf{Critique:} Is the complexity of the alpha expression plausibly justified by an inferred economic rationale, or does it seem arbitrary/excessive, suggesting fitting to noise?
                \end{itemize}
            \item \textbf{Principled Development vs. Data Dredging:}
                \begin{itemize}
                    \item[] \textbf{Critique:} Does the refinement history indicate hypothesis-driven improvements, or does it suggest excessive optimization and curve-fitting (e.g., frequent, unjustified parameter tweaks)?
                \end{itemize}
            \item \textbf{Transparency vs. Opacity:}
                \begin{itemize}
                    \item[] \textbf{Critique:} Is the alpha's logic reasonably interpretable despite its complexity, or is it opaque, potentially masking overfitting?
                \end{itemize}
        \end{enumerate}

        \vspace{1ex}
        \noindent\textbf{Scoring \& Output:}
        \begin{itemize}
            \item Assign a single \textbf{Overfitting Risk Score} from 0 to 10.
                \begin{itemize}
                    \item \textbf{10 = Very Low Risk} (High confidence in generalization)
                    \item \textbf{0 = Very High Risk} (Low confidence in generalization)
                \end{itemize}
            \item Use the full 0-10 range to differentiate risk levels effectively.
            \item Provide a concise, one-sentence \textbf{Justification} explaining the score, citing the key factors from the criteria.
            \item Format the output as JSON, like the examples below:
        \end{itemize}

        \vspace{1ex}
        \noindent\textbf{Example JSON Outputs:}

\begin{lstlisting}
{
    "reason": "Complexity is justified by a strong rationale; principled refinement history suggests low risk.",
    "score": 9
}
\end{lstlisting}

\begin{lstlisting}
{
    "reason": "Plausible rationale, but some expression opacity and parameter tuning in history indicate moderate risk.",
    "score": 5
}
\end{lstlisting}

\begin{lstlisting}
{
    "reason": "High risk inferred from opaque expression lacking clear rationale, supported by history showing excessive tuning.",
    "score": 1
}
\end{lstlisting}
        \vspace{1ex}
        \hrulefill
    \end{minipage}
\caption{Alpha Overfitting Risk Assessment Prompt}
\label{fig:overfitting_assessment_prompt}
\end{figure}

\begin{figure}[htbp]
    \centering
    \small
    \begin{minipage}{0.95\textwidth}
        \hrulefill
        \vspace{1ex}

        \noindent\textbf{Task Description:}

        \noindent There is an alpha factor used in quantitative investment to predict asset price trends.
        Please improve it according to the following suggestions and provide the improved alpha expression.

        \vspace{1ex}
        \noindent\textbf{Available Data Fields:}

        \noindent The following data fields are available for use: \texttt{\{available\_fields\}}

        \vspace{1ex}
        \noindent\textbf{Available Operators:}

        \noindent The following operators are available for use:\texttt{\{available\_operators\}}

        \vspace{1ex}
        \noindent\textbf{Alpha Suggestions:}
        \begin{enumerate}
            \item The alpha value should be dimensionless (unitless).
            \item All look-back windows and other numerical parameters used in the alpha calculation MUST be represented as named parameters in the pseudo-code. These parameter names MUST follow Python naming conventions (e.g., \texttt{lookback\_period}, \texttt{volatility\_window}, \texttt{smoothing\_factor}).
            \item The alpha should have NO MORE than 3 parameters in total.
            \item The pseudo-code should represent the alpha calculation step-by-step, using only the "Available Operators" and clearly defined parameters. Each line in the pseudo-code should represent a single operation.
            \item Use descriptive variable names in the pseudo-code that clearly indicate the data they represent.
            \item When designing alpha expressions, try to avoid including the following sub-expressions: \texttt{\{freq\_subtrees\}}
        \end{enumerate} 

        \vspace{1ex}
        \noindent\textbf{Original alpha expression:}
\begin{lstlisting}
{origin_alpha_formula}
\end{lstlisting}

        \vspace{1ex}
        \noindent\textbf{Refinement suggestions:}
        
        \noindent NOTE: The following improvement suggestions do not need to be all adopted; they just need to be considered and reasonable ones selected for adoption.
\begin{lstlisting}
{refinement_suggestions}
\end{lstlisting}

        \vspace{1ex}
        \noindent\textbf{Formatting Requirements:}
        
        \noindent The output must be in JSON format with three key-value pairs:
        \begin{enumerate}
            \item \textbf{"name"}: A short, descriptive name for the alpha (following Python variable naming style, e.g., \texttt{price\_volatility\_ratio}).
            \item \textbf{"description"}: A concise explanation of the alpha's purpose or what it measures. Avoid overly technical language. Focus on the intuition behind the alpha.
            \item \textbf{"pseudo\_code"}: A list of strings, where each string is a line of simplified pseudo-code representing a single operation in the alpha calculation. Each line should follow the format: \texttt{variable\_name = op\_name(input=[input1, input2, ...], param=[param1, param2, ...])}, where:
                \begin{itemize}
                    \item \texttt{variable\_name} is the output variable of the operation.
                    \item \texttt{op\_name} is the name of one of the "Available Operators".
                    \item \texttt{input1}, \texttt{input2}, ... are input variables (either from "Available Data Fields" or previously calculated variables, cannot be of a numeric type).
                    \item \texttt{param1}, \texttt{param2}, ... are parameter names defined in the alpha requirements.
                \end{itemize}
        \end{enumerate}

        \vspace{1ex}
        \noindent The format example is as follows:
\begin{lstlisting}
{
    "name": "volatility_adjusted_momentum",
    "description": "......",
    "pseudo_code": [......]
}
\end{lstlisting}
        \vspace{1ex}
        \hrulefill
    \end{minipage}
\caption{Alpha Refinement Prompt}
\label{fig:refinement_prompt}
\end{figure}